\documentclass[10pt,twocolumn,letterpaper]{article}

\usepackage{wacv}
\makeatletter
\@namedef{ver@everyshi.sty}{}
\makeatother
\usepackage{times}
\usepackage{epsfig}
\usepackage{graphicx}
\usepackage{amsmath}
\usepackage{amssymb}
\usepackage{booktabs}

\usepackage[accsupp]{axessibility}  
\usepackage{caption}
\usepackage{subcaption}
\usepackage{color}
\usepackage[font=small,labelfont=bf]{caption}
\usepackage{xspace}
\usepackage{xcolor,colortbl}
\newcommand{\DFOSDA}{\texttt{D$^2$F2WOD}\xspace}

\usepackage{multirow}
\usepackage{mathtools,tikz}
\DeclareRobustCommand\sampleline[1]{%
  \tikz\draw[#1] (0,0) (0,\the\dimexpr\fontdimen22\textfont2\relax)
  -- (2em,\the\dimexpr\fontdimen22\textfont2\relax);%
}

\usepackage{booktabs}
\usepackage{soul}
\graphicspath{ {./images/} }
\DeclareMathOperator*{\argmax}{argmax}
\usepackage{mathtools,tikz}
\DeclareRobustCommand\sampleline[1]{%
  \tikz\draw[#1] (0,0) (0,\the\dimexpr\fontdimen22\textfont2\relax)
  -- (2em,\the\dimexpr\fontdimen22\textfont2\relax);%
}

\usepackage[pagebackref=true,breaklinks=true,colorlinks,bookmarks=false]{hyperref}
\usepackage[capitalize]{cleveref}
\crefname{section}{Sec.}{Secs.}
\Crefname{section}{Section}{Sections}
\Crefname{table}{Table}{Tables}
\crefname{table}{Tab.}{Tabs.}

\makeatletter
\DeclareRobustCommand\onedot{\futurelet\@let@token\@onedot}
\def\@onedot{\ifx\@let@token.\else.\null\fi\xspace}

\def\eg{\emph{e.g}\onedot} 
\def\ie{\emph{i.e}\onedot}

\makeatother

\def\wacvPaperID{} 

\wacvalgorithmstrack   

\wacvfinalcopy 

\ifwacvfinal
\usepackage[breaklinks=true,bookmarks=false]{hyperref}
\else
\usepackage[pagebackref=true,breaklinks=true,colorlinks,bookmarks=false]{hyperref}
\fi

\begin{document}

\title{\DFOSDA: Learning Object Proposals for Weakly-Supervised Object Detection via Progressive Domain Adaptation}

\author{Yuting Wang\\
Rutgers University\\
Piscataway, NJ\\
{\tt\small yw632@cs.rutgers.edu}
\and
Ricardo Guerrero\\
Samsung AI Center\\
Cambridge, UK\\
{\tt\small r.guerrero@samsung.com}
\and
Vladimir Pavlovic\\
Rutgers University\\
Piscataway, NJ\\
{\tt\small vladimir@cs.rutgers.edu}
}

\maketitle
\thispagestyle{empty}

\begin{abstract}
\label{sec:abs}
Weakly-supervised object detection (WSOD) models attempt to leverage image-level annotations in lieu of accurate but costly-to-obtain object localization labels. This oftentimes leads to substandard object detection and localization at inference time. To tackle this issue, we propose \DFOSDA, a \textbf{D}ual-\textbf{D}omain \textbf{F}ully-to-\textbf{W}eakly Supervised \textbf{O}bject \textbf{D}etection framework that leverages synthetic data, annotated with precise object localization, to supplement a natural image target domain, where only image-level labels are available. In its warm-up domain adaptation stage, the model learns a fully-supervised object detector (FSOD) to improve the precision of the object proposals in the target domain, and at the same time learns target-domain-specific and detection-aware proposal features. In its main WSOD stage, a WSOD model is specifically tuned to the target domain. The feature extractor and the object proposal generator of the WSOD model are built upon the fine-tuned FSOD model. 
We test \DFOSDA on five dual-domain image benchmarks. The results show that our method results in consistently improved object detection and localization compared with state-of-the-art methods.
\end{abstract}

\section{Introduction}
\label{sec:intro}
Object detection has achieved remarkable progress over the past few years, mostly through the development of deep neural network architectures~\cite{ren2015faster,carion2020end}. However, training such deep neural networks needs large amounts of manually annotated images. Obtaining these annotations is costly and time-consuming. Thus, reducing these costs is of great importance, and many weakly-supervised object detection (WSOD) methods~\cite{bilen2016weakly,tang2017multiple,tang2018pcl} have been developed accordingly. 
WSOD methods alleviate the reliance on precise object localization information by training detection architectures using only {\em image-level} annotations.

\begin{figure}[t]
  \centering
  \includegraphics[trim={0cm +10cm 0cm +0cm},clip,scale=0.25]{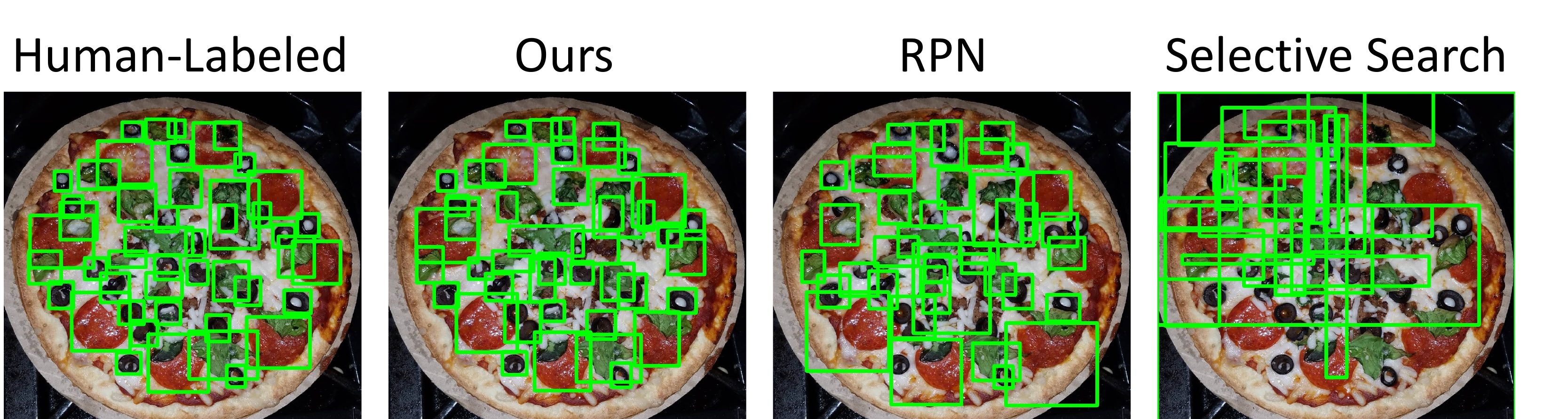}
   \caption{Illustration of human-labeled objects contrasted to object proposals generated by our $\DFOSDA_\text{warm-up}$, RPN of Faster R-CNN trained on synthetic data alone, and Selective Search (SS), on the RealPizza10 dataset. It demonstrates the benefit of our learned object proposal generator (warm-up stage) over SS. SS often fails to generate accurate bounding boxes, making it hard to improve classification accuracy. It also shows that our $\DFOSDA_\text{warm-up}$ is better than learned RPN of Faster R-CNN trained on synthetic data alone. Our warm-up domain adaptation stage can improve the precision of the object proposals in the target domain.
   }
   \label{fig:proposal_all}
\end{figure}

Most existing WSOD algorithms~\cite{bilen2016weakly,tang2017multiple,tang2018pcl,wan2019c,ren2020instance,gao2019c,huang2020comprehensive} are based on multiple instance learning (MIL)~\cite{dietterich1997solving}. They treat images as bags of object proposals, which are produced by an object proposal generator~\cite{uijlings2013selective,zitnick2014edge}. Although many promising results have been achieved by WSOD, they are still not
comparable to fully-supervised object detectors (FSOD)~\cite{ren2015faster,carion2020end}. 
One of the main reasons is that state-of-the-art object proposal generators still cannot produce accurate object proposals --  this is a particularly serious issue for {\em in-the-wild} images with multiple complex non-rigid objects and cluttered background, as shown in~\autoref{fig:proposal_all}.

To overcome this difficulty, we introduce a simple 
object proposal generation strategy that can be applied to different WSODs to improve their detection performance. Our key insight is to cast WSOD as a {\em domain adaptation problem} -- while target ``natural'' images often lack localization labels, localization is ``freely'' available for ``non-photographic'' synthetic images. 
For instance, when synthesizing images such as  SyntheticPizza10~\cite{papadopoulos2019make}, localization and identity labels are available as a byproduct of the generation process. Highly stylized images (\eg, Clipart1K~\cite{inoue2018cross}, Watercolor2K~\cite{inoue2018cross}, and Comic2K~\cite{inoue2018cross}) are likewise easier to annotate than natural images, where objects may exhibit complex changes in share or appearance. 
In this work, we are interested in leveraging 
fully-annotated non-photographic datasets to support accurate object 
localization in real-world datasets. To this end, we propose a 
\textbf{D}ual-\textbf{D}omain \textbf{F}ully-to-\textbf{W}eakly Supervised \textbf{O}bject \textbf{D}etection (\DFOSDA) framework, which is able to produce accurate object proposals using
image-level labels of natural images along with fully-supervised non-photographic images through \textit{progressive domain adaptation}
of
an FSOD model.

Given the large domain gap between the source and the target, across both foreground and background (F\&B), it is critical to
(1) individually address the adaptation of F\&B in a disentangled manner when feasible, and (2) reduce the domain gap in a gradual manner to control the propagation of errors.
In our work, we progressively adapt an FSOD model from source images to the target domain in five steps. 
First, we build an initial bridge between the non-photographic source and the real-world target domains using unpaired image-to-image translation (I2I), such as~\cite{zhu2017unpaired}.
This creates ``target-like'' intermediate images with location-accurate object instances but divergent appearance.
Instead of the common practice of initializing an FSOD on this intermediate domain, we further reduce the domain gap by employing a copy-paste augmentation technique sourced in~\cite{yun2019cutmix,ghiasi2021simple}
to fuse the translated object appearance with 
real background images and create a second transfer-labeled intermediate domain. This domain serves as the preliminary stage for initializing an FSOD, to be used for pseudo labeling (PL)~\cite{lee2013pseudo} in the subsequent WSOD learning phase on the real target domain.  
However, the typical number of confident pseudo-labeled instances resulting from the initialized FSOD and needed for WSOD is insufficient for effective adaptation to the target domain. To that end, we re-employ the previously used augmentation technique to increase the number of confident PL instances.
Finally, we learn a detection head utilizing these target-like object proposal features. Our \DFOSDA achieves consistent improvements compared with the state-of-the-art methods, 
offering a strong baseline for WSOD models.

\textbf{Our contributions} are three-fold: (1) We propose a framework for object proposal generation based on domain adaptation, applicable to different WSODs, including OICR~\cite{tang2017multiple}, and CASD~\cite{huang2020comprehensive}. The five-step progressive domain adaptation process exploits gradual adaptation of the FSOD on generated samples, as well as with decoupled focus on foreground and background, and it can be seamlessly integrated with different types of FSOD backbones such as Faster R-CNN~\cite{ren2015faster} and transformer-based detectors~\cite{carion2020end}.
(2) We construct a dual-domain image benchmark SyntheticPizza10 $\rightarrow$ RealPizza10 with non-photographic images as the source and real-world images as the target domains.
(3) The experimental results show that our \DFOSDA achieves state-of-the-art performance on five benchmarks.
\section{Related Work}
\label{sec:related work}
\noindent\textbf{Weakly-Supervised Object Detection.}
WSOD methods generally aim to exploit only image-level annotations, as opposed to the fine-grained object localization usually used in FSOD.
Existing methods mainly cast WSOD as a multiple-instance learning (MIL) problem, where objects are not necessarily centered in images and there is cluttered background~\cite{nguyen2009weakly}. 
In MIL-based models, an image is interpreted as a bag of potential object instances. These models generally consist of three components: feature extractor (FE), object proposal (OP) generator, and detection head (DH). Given an image, they first feed it into the OP generator and 
the 
FE to generate proposals and features maps, respectively. Then, the feature maps and object proposals (OPs) are forwarded into a Spatial Pyramid Pooling (SPP) layer~\cite{van2017learning} or a Region-of-Interest (RoI) pooling layer~\cite{ren2015faster} to produce fixed-size object proposal features. Finally, these feature vectors are fed into the DH to classify and localize objects.
End-to-end weakly-supervised deep detection network (WSDDN)~\cite{bilen2016weakly} proposes one of the first MIL frameworks. Based on Fast R-CNN~\cite{girshick2015fast}, it introduces a two-stream network to perform classification and localization, respectively.
However, in WSDNN, the top ranking OPs may only cover the most discriminative parts of the objects instead of whole object instances, due to a lack of supervision in terms of precise localization information in the training process. 
Subsequent work~\cite{tang2017multiple,tang2018pcl,wan2019c,arun2019dissimilarity,zeng2019wsod2,ren2020instance,gao2019c,huang2020comprehensive,tang2018weakly} aims to alleviate
this problem by extending WSDDN.
One of the key factors that affect the performance of WSOD is the quality of OPs. Many existing methods are built upon unsupervised RoI extraction, such as selective search (SS)~\cite{uijlings2013selective} and edge boxes (EB)~\cite{zitnick2014edge}.
To generate OPs, SS uses both exhaustive search and segmentation, and EB uses object edges. 
~\cite{zhang2021hierarchical} proposes a hierarchical region proposal refinement network and ~\cite{tang2018weakly} proposes a two-stage region proposal network, to refine proposals gradually. Some other work, such as W2N~\cite{huang2022w2n}, continues refine the noisy dataset generated by a well-trained WSOD with semi-supervised learning.

Different from the above methods, in this work, we first cast WSOD as a domain adaptation problem, by leveraging an auxiliary source domain to pre-train an FSOD model. The FSOD model is progressively adapted from the source to the target domains. After we obtain the adapted FSOD model, we treat it as the weakly-supervised OP generator in the WSOD settings, and at the same time the FE of the FSOD is treated as the pre-trained FE for the WSOD model.

\noindent\textbf{Domain Adaptation for Object Detection.}
Domain adaptation typically involves two domains, namely source and target domains. Most of existing domain adaptation methods aim to address the domain shift between a fully-labeled source domain and an unlabeled or weakly-labeled target domain, which is formulated as unsupervised or weakly-supervised domain adaptation, respectively. 
State-of-the-art domain adaptation for object detection introduces different strategies to reduce the domain divergence. For example, adversarial feature learning is leveraged to adapt object detectors to a target domain with the help of a domain discriminator~\cite{chen2018domain,saito2019strong,sindagi2020prior,hsu2020progressive,vs2021mega}, thus producing domain invariant features. Highly confident predictions generated by a source detector are used as pseudo-labels to fine-tune the detector on the target domain~\cite{inoue2018cross,kim2019self,zhao2020collaborative,roychowdhury2019automatic}. Similarly, an unpaired 
I2I model~\cite{inoue2018cross,rodriguez2019domain,hsu2020progressive} can be employed to map a source image to a target-like image. Introducing this target-like
domain mitigates the difficulty of direct transfer between source and target with a large domain gap.

Different from the aforementioned approaches, our method decouples the domain shift into the foreground and background shift. This makes it possible to gradually, in a focused manner, adapt the detector from source to target. We also use data augmentation in the adaptation stage, since augmentations such as color jittering~\cite{szegedy2015going}, mixup~\cite{zhang2019bag} and copy-paste~\cite{yun2019cutmix,ghiasi2021simple} can have major impact on image classification and object detection. Furthermore, OPs generated by the adapted object detector are augmented by an additional refinement of proposal branches using the detection heads in the WSOD settings. This refinement improves the network's ability to classify and localize the OPs.

\section{Methodology}
The proposed
Dual-Domain Fully-to-Weakly Supervised Object Detection framework (\DFOSDA) aims to address the lack of object localization information in the target domain by formulating WSOD as a domain adaptation problem. It decouples WSOD model training into two stages -- 
domain adaptation and WSOD. 
In the domain adaptation stage, 
we {\em progressively} learn a domain-adaptive FSOD by leveraging an auxiliary source domain as warm-up.
In the WSOD stage, this adapted FSOD is used to initialize the WSOD model, which is then refined on the target domain. \DFOSDA is a general framework that can employ different FSOD and WSOD methods. Here, we focus on two representative FSOD backbones -- 
Faster R-CNN~\cite{ren2015faster} and 
DETR~\cite{carion2020end},
and two representative WSOD models -- the widely-used OICR~\cite{tang2017multiple} and the state-of-the-art CASD~\cite{huang2020comprehensive}. In this section, we first formulate the problem, followed by the framework overview, the details of the architecture, and the training procedure for each stage of \DFOSDA.

\subsection{Problem Formulation}
\autoref{fig:architecture} illustrates our \DFOSDA approach.  Our goal is to detect object instances in a real-world, weakly-supervised target domain $\mathcal{T}$ (\eg, real pizza in~\autoref{fig:architecture}) by leveraging 
a
non-photographic source domain $\mathcal{S}$ (\eg, synthetic pizza in~\autoref{fig:architecture}). For this problem, we have access to images with only image-level annotations (\ie, class labels) in $\mathcal{T}$ and images with rich instance-level annotations (\ie, class labels and bounding boxes) in $\mathcal{S}$.

Formally, $\mathbf{X}_s\in{\mathbb{R}}^{h{\times}w{\times}3}$ denotes an RGB image from $\mathcal{S}$, where $h$ and $w$ are the height and width of the image, respectively.
$\mathbf{Y}^{(f)}_s = \{(\mathbf{b}_{1},c_{1}),\ldots,(\mathbf{b}_{N_s},c_{N_s})\}$ indicates the instance-level full-annotation associated with $\mathbf{X}_s$, where $\mathbf{b}_{i}\in{\mathbb{R}}^{4}$ is the $i$-th object localization bounding box defined by $[x_\mathrm{min},y_\mathrm{min},x_\mathrm{max},y_\mathrm{max}]$ that specifies its top-left corner $(x_\mathrm{min},y_\mathrm{min})$ and its bottom-right corner $(x_\mathrm{max},y_\mathrm{max})$,
and $c_{i} \in \{1,\ldots,C \}$ is its category label.
$N_s$ is the number of object instances associated with $\mathbf{X}_s$.
The classes to be detected in  $\mathcal{T}$ are shared with $\mathcal{S}$, and $C$ is the number of object categories in the two domains. Similarly,  $\mathbf{X}_t\in{\mathbb{R}}^{h{\times}w{\times}3}$ denotes an RGB image from $\mathcal{T}$, and $\mathbf{Y}^{(w)}_t = [y_1,\ldots,y_C]\in\{0,1\}^C$ denotes the image-level weak-supervision, where $y_c \in \{0,1\}$ indicates the absence (presence) of at least one instance of $c$-th category. $N_t$ is the number of present object classes associated with $\mathbf{X}_t$. We denote $\mathbf{V}_{j}$ as object proposal feature vectors of images from domain 
$j\in\{\mathcal{S},\mathcal{T}\}$. 
In this work, we aim to learn an object detector for the target domain, $\hat{\mathbf{Y}}^{(f)} = f(\mathbf{X}|\boldsymbol{\theta}), \mathbf{X} \in \mathcal{T}, $ by leveraging both the fully 
annotated data
$\mathcal{D}_s = \{ (\mathbf{Y}_s^{(f)}, \mathbf{X}_s) \}$ from $\mathcal{S}$ and the weakly annotated data $\mathcal{D}_t = \{ (\mathbf{Y}_t^{(w)}, \mathbf{X}_t) \}$ from $\mathcal{T}$; in other words, $\boldsymbol{\theta}^* \leftarrow \mathcal{D} = \{ \mathcal{D}_s \cup \mathcal{D}_t \}$.

\subsection{Approach Overview}
\begin{figure*}[ht!]
\centering
\includegraphics[width=\linewidth]{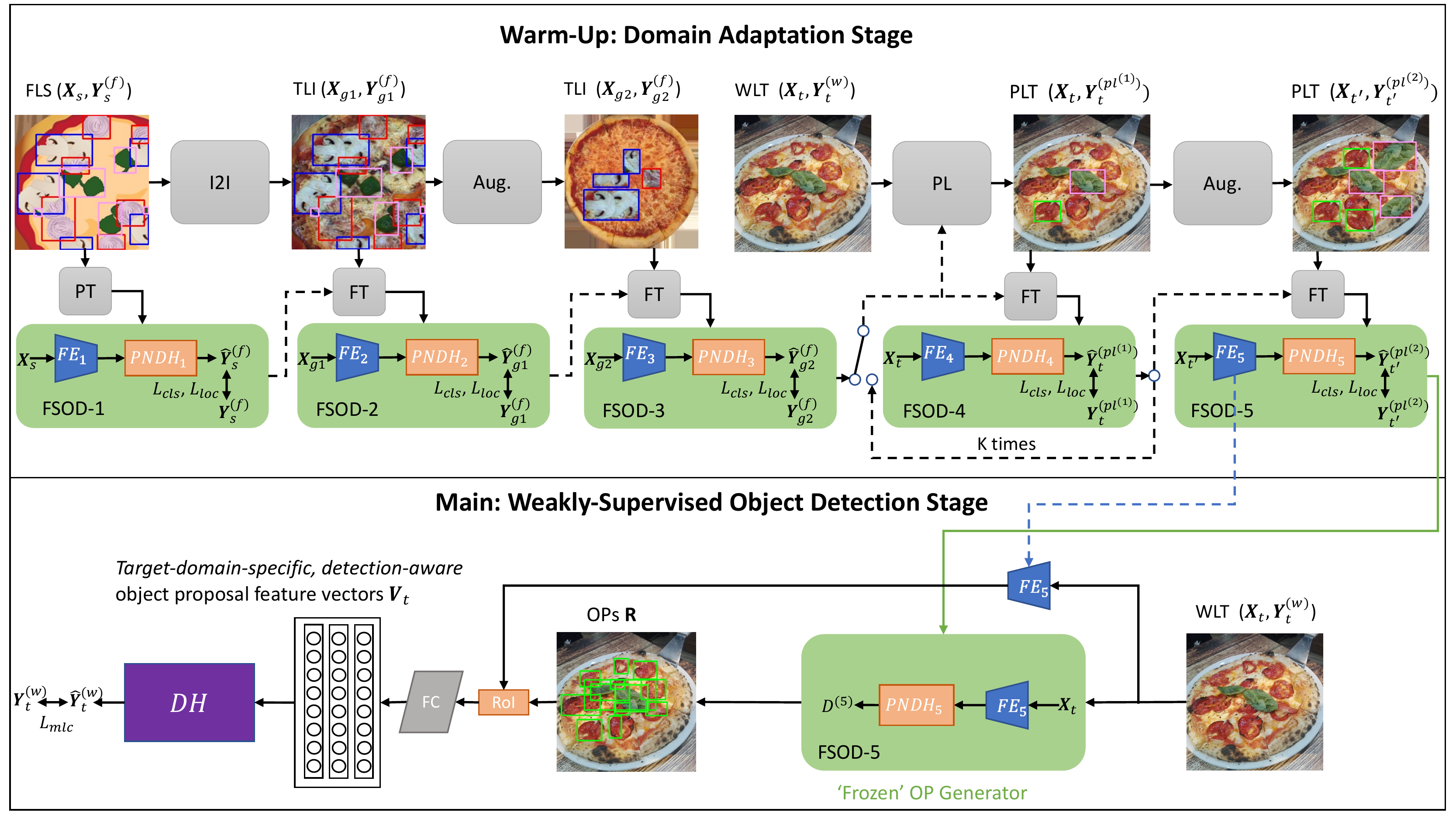}
\caption{An overview of our Dual-Domain Fully-to-Weakly
Supervised Object Detection Architecture (\DFOSDA). \textbf{Upper block:} warm-up domain adaptation stage; \textbf{lower block:} main weakly-supervised object detection stage. 
Here 
`A\sampleline{dashed}B' 
denotes that the parameters of module B are initialized from module A's parameters. 
`A\sampleline{}B' 
denotes that the output of module A is used as an input of module B or module B directly copies module A's parameters without further fine-tuning. `PNDH' denotes the proposal networks and detection heads in an FSOD model.
}
\label{fig:architecture}
\end{figure*}
To boost the performance of a WSOD model on $\mathcal{T}$, our key insight is to jointly improve the precision of the OPs, and learn {\em target-domain-specific} and {\em detection-aware} proposal features. 
To this end, our \DFOSDA exploits the fully-labeled $\mathcal{S}$ (FLS) domain 
and introduces a dual-stage training scheme as shown in~\autoref{fig:architecture}. In the {\em warm-up} domain adaptation stage, an FSOD is pre-trained (PT) on $\mathcal{S}$ and progressively fine-tuned (FT) on (1) 
a transfer-labeled intermediate (TLI) domain $\mathcal{G}_1$, 
(2) an augmented (Aug.) transfer-labeled intermediate domain $\mathcal{G}_2$, and 
then on (3) the pseudo-labeled target (PLT) domain $\mathcal{T}$ and (4) the augmented pseudo-labeled target domain $\mathcal{T}$. 
As shown in~\autoref{fig:architecture}, $\mathcal{G}_1$ is constructed as target-like instances with accurate transferred localization information, and $\mathcal{G}_2$ is constructed as target-like images with accurate transferred localization information and real background, thus bridging the $\mathcal{S}$ and $\mathcal{T}$ and facilitating the adaptation. In the {\em main} WSOD stage, an MIL-based WSOD model is specifically tuned to $\mathcal{T}$. The FE and the OP generator of the WSOD model are built upon the fine-tuned FSOD model.

\subsection{Warm-Up Domain Adaptation Stage: Learning Domain-Specific Features \& Object Proposals}
\label{sec:main_warmup}
The warm-up stage of \DFOSDA trains an FSOD model on the dual $\mathcal{S}$-$\mathcal{T}$, which provides the pre-trained deep FE and OP generator to produce object proposal feature vectors on $\mathcal{T}$. This FSOD model is later used in the main stage for initializing the WSOD model. Our method generalizes across different FSODs. 
Here,
we adopt two 
architectures for the FSOD model -- 
Faster R-CNN~\cite{ren2015faster} and 
Sparse DETR~\cite{roh2021sparse} from the DETR~\cite{carion2020end} family of object detectors\footnote{
Sparse DETR enhanced the efficiency of DETR and improved the performance on small objects datasets, and thus we choose Sparse DETR here.
}. 

\subsubsection{Progressive Domain Adaptation.}
Directly training the FSOD model on the dual domain is challenging, because of (1) substantial data distribution shift between source (non-photographic images) and target (natural images) domains in both foreground and background, and (2) significant supervision discrepancy between source (fully-labeled) and target (lack of localization information) domains. Inspired by DT+PL~\cite{inoue2018cross}, we overcome this difficulty by generating an intermediate domain $\mathcal{G}_{1}$ and $\mathcal{G}_{2}$ with instance-level annotations transferred from the source domain. Correspondingly, we introduce a five-step progressive domain adaptation strategy (upper part in \autoref{fig:architecture}) that first pre-trains the FSOD-1 model on the fully-labeled $\mathcal{S}$, and gradually fine-tunes it on the transfer-labeled $\mathcal{G}_1$ and augmented transfer-labeled $\mathcal{G}_2$ to be FSOD-2 and FSOD-3, and then on the first-round pseudo-labeled $\mathcal{T}$ and second-round augmented pseudo-labeled $\mathcal{T}$ to be FSOD-4 and FSOD-5.

\noindent\textbf{Automated generation of intermediate domains for initial adaptation.} 
To facilitate the adaptation, a desired property of the intermediate domain should be that {\em its images are similar to the target images while having accurate localization information}. To this end, we generate the intermediate domain images as composition of photo-realistic, target-like objects guided by the layout of objects in the source images, thus allowing direct transfer of localization annotations from the source images to the generated images. 

Specifically, since there are no corresponding image pairs between $\mathcal{S}$ and $\mathcal{T}$ domains, we train CycleGAN~\cite{zhu2017unpaired}, an unpaired
I2I network, to map source images $\mathbf{X}_s$ to domain $\mathcal{G}_1$ intermediate to the target $\mathcal{T}$:
\begin{equation}
\mathbf{X}_{g_1} = f_{\mathcal{S}\rightarrow\mathcal{G}_1}( \mathbf{X}_s ),
\end{equation} 
where $\mathbf{X}_{g_1}\in{\mathbb{R}}^{h{\times}w{\times}3}$ is the image generated by I2I network.
Given this I2I mapping, we transfer the labels from instances in $\mathbf{X}_s$ to those in $\mathbf{X}_{g_1}$ as
\begin{equation}
\mathbf{Y}_{g_1}^{(f)} = \mathbf{Y}_s^{(f)}: \quad \mathbf{X}_{g_1} = f_{\mathcal{S}\rightarrow\mathcal{G}_1}( \mathbf{X}_s ).
\end{equation}
Using the intermediate images $\mathbf{X}_{g_1}$ together with their instance-level annotations $\mathbf{Y}^{(f)}_{g_1}$, we fine-tune the FSOD-1 model, pre-trained on $\mathcal{S}$, into FSOD-2.

To make $\mathbf{X}_{g_1}$ closer to the $\mathbf{X}_{t}$ images, we focus on separately bridging the foreground and background gap.  
Specifically, we employ an object-aware data augmentation based on copy-paste~\cite{yun2019cutmix} to map $\mathbf{X}_{g_1}$ to $\mathbf{X}_{g_2}$ images. For each $\mathbf{X}_{g_1}$ image, we randomly copy several foreground object instances from $\mathcal{G}_1$, with resizing and flipping transformations, and paste them onto the real-world target background images from $\mathcal{T}$ to generate $\mathbf{X}_{g_2}$. 
Using the augmented intermediate images $\mathbf{X}_{g_2}$ together with their instance-level annotations $\mathbf{Y}^{(f)}_{g_2}$, we fine-tune the FSOD-2 model to FSOD-3.

\noindent\textbf{Instance-level pseudo-annotation of target images for continual adaptation.}
While the intermediate domains $\mathcal{G}_1$ and $\mathcal{G}_2$
partly
bridge the source and target domains, there is still non-negligible domain shift between the intermediate and target domains. For example, the synthesized objects translated via CycleGAN are still different from those in the target images; the layout of objects in the intermediate domain is restrictive to that in $\mathcal{S}$ and lacks the real-world variation in $\mathcal{T}$. Therefore, to achieve good detection performance on the target domain, we need to further fine-tune the FSOD-3 model on the target domain $\mathcal{T}$ as FSOD-4. For this purpose, we use FSOD-3, initially fine-tuned on $\mathcal{G}_2$, to produce instance-level pseudo-annotations $\mathbf{Y}^{(pl^{(1)})}_t = \{(\mathbf{b}_1,c_1),\ldots,(\mathbf{b}_{N_t},c_{N_t})\}$ for each weakly-labeled target (WLT) image~$\mathbf{X}_t$. 

Specifically, for each image $\mathbf{X}_t$, we first obtain the predictions $D^{(3)}$ from the 
FSOD-3 model:
\begin{equation}
D^{(3)} = \{D_1, ..., D_{C}\} = f_{\text{FSOD-3}}(\mathbf{X}_t),
\end{equation}
where $D_j$ indicates 
all 
predictions belonging to class $j\in\{1,\ldots,C\}$. $D_j = \{d_1, ..., d_{N_j}\}$, $N_j$ is the number of 
class $j$ detections, 
$d_m = (p_{m},\hat{\mathbf{b}}_{m},j)$, 
and 
$p_m \in \mathbb{R}$ indicates the probability of detection $\hat{\mathbf{b}}_{m}$ belonging to class $j$. 
For each ground-truth object class $c$, we select the top-1 confident prediction $d_q$ from $D_c$, and we add $(\hat{\mathbf{b}}_{q},c)$ to $\mathbf{Y}^{(pl^{(1)})}_t$:
\begin{equation}
d_q= (p_{q},\hat{\mathbf{b}}_{q},c):\quad y_{c} = 1, \quad q = \argmax_{m} p_{m}.
\end{equation}
The FSOD-3 model is subsequently fine-tuned on the target images $\mathbf{X}_t$ with instance-level pseudo-annotations $\mathbf{Y}^{(pl^{(1)})}_t$ into FSOD-4, finally adapting from the target-like $\mathcal{G}_2$ to $\mathcal{T}$. In principle, it can be performed $K$ times to generate instance-level pseudo-annotations $\mathbf{Y}^{(pl^{(k)})}_t$ and adapted into FSOD-($3+k$), where $k \in \{1,\ldots,K \}$.

However, the typical number of confident pseudo-labeled instances resulting from the FSOD-3 is insufficient for effective adaptation to the target domain. To add instances annotations, we use the copy-paste augmentations again to produce those object instances. We repeat the previous step, in which the FSOD-4 model is used to produce instance-level pseudo-annotations $\mathbf{Y}^{(pl^{(2)})}_t$. For each pseudo-labeled instance  $(\hat{\mathbf{b}}_{q},c)$ in $\mathbf{X}_t$, we copy and randomly paste it $L$ times as $\{(\hat{\mathbf{b}}_{q_1},c)...,(\hat{\mathbf{b}}_{q_L},c)\}$ onto the original target image $\mathbf{X}_t$ and produce an augmented image $\mathbf{X}^{'}_{t}$ with new pseudo-annotations ${\mathbf{Y}^{(pl^{(2)})}_{t}}$.
The FSOD-4 is subsequently fine-tuned on the augmented targets $\mathbf{X}^{'}_{t}$ with instance-level pseudo-annotations ${\mathbf{Y}^{(pl^{(2)})}_{t}}$ into FSOD-5, thus adapting from $\mathcal{T}$ to the augmented $\mathcal{T}$.

\subsection{Main WSOD Stage: Classification and Localization Refinement of Object Proposals}
\label{sec:main_wsod}
In the main stage of \DFOSDA, we exploit the FSOD-5 model obtained in the warm-up stage to initialize a WSOD model and train it on the real-world target data. As explained in \autoref{sec:related work}, an MIL-based WSOD model consists of a FE, an OP generator, and a DH. We initialize the FE of the WSOD model with the FE of the fine-tuned FSOD-5 (blue block in \autoref{fig:architecture}) and continually train it on $\mathcal{T}$. We replace the standard selective search based OP generator of the WSOD model by the entire fine-tuned FSOD-5 (green block in \autoref{fig:architecture}), which is not trained in the WSOD training procedure. Note that here we treat the detection output of the FSOD-5 as the object proposals of the WSOD. This strategy can be seamlessly applied to different types of WSODs, and here we consider the widely-used OICR~\cite{tang2017multiple} and the state-of-the-art CASD~\cite{huang2020comprehensive}. By doing so, our 
proposed
model significantly outperforms existing WSOD methods due to: (1) target-domain-specific pre-trained features, (2) detection-aware pre-trained features, and (3) target-domain-specific object proposals.  

\noindent\textbf{Generating object proposals and its features.}
Given an image $\mathbf{X}_t$, the OP generator aims to obtain $M_t$ bounding boxes $\mathbf{R} = \{\textbf{b}_1, ..., \textbf{b}_{M_t}\}$ associated with $\mathbf{X}_t$. 
To this end, for each image $\mathbf{X}_t$, we first obtain the predictions $D^{(5)}$ from the fine-tuned FSOD-5 model:
\begin{equation}
D^{(5)} = \{D_1, ..., D_{C}\} = f_{\text{FSOD-5}}(\mathbf{X}_t).
\end{equation}
Given that the number of predictions produced by DETR is much less than that of Faster R-CNN, we adopt different proposal generation strategies. For DETR, all predicted bounding boxes are added to $\mathbf{R}$. For Faster R-CNN, we select predicted bounding boxes $\hat{\mathbf{b}}_{m}$ belonging to the ground-truth classes to $\mathbf{R}$. 
Using the FE followed by an RoI pooling layer and two fully-connected (FC) layers for the WSOD model, we then obtain $d$-dimensional object proposal feature vectors $\mathbf{V}_t\in{\mathbb{R}}^{{d}{\times}{M_t}}$ for each input image $\mathbf{X}_t$ (lower part in \autoref{fig:architecture}).

\noindent\textbf{Classification and localization refinement of object proposals.}
These object proposal feature vectors $\mathbf{V}_t$ are fed into the detection head of OICR~\cite{tang2017multiple} or CASD~\cite{huang2020comprehensive} to classify and
localize objects. Please refer to \autoref{sec:wsod_detail} for the details.

\section{Experimental Results}

\noindent\textbf{Benchmarks.} We evaluate our method on five dual-domain image benchmark pairs: SyntheticPizza10~\cite{papadopoulos2019make} $\rightarrow$ RealPizza10~\cite{papadopoulos2019make}, Clipart1K~\cite{inoue2018cross} $\rightarrow$ VOC2007~\cite{everingham2010pascal}, Watercolor2K~\cite{inoue2018cross} $\rightarrow$ VOC2007-sub, Comic2K~\cite{inoue2018cross} $\rightarrow$ VOC2007-sub, and Clipart1K $\rightarrow$ MS-COCO-sub~\cite{lin2014microsoft} datasets. We construct the SyntheticPizza10 dataset from~\cite{papadopoulos2019make} by including single-layer images and removing the pizza base-only images (\ie, without any toppings). 
RealPizza10 is a subset of the PizzaGAN~\cite{papadopoulos2019make}, containing 9,213 real images annotated with 13 toppings. As we use
pseudo-labeling, we require the classes in ($\mathcal{S}$, $\mathcal{T}$) to be the same. Thus, we remove images from the PizzaGAN dataset having only spinach, arugula, or corn, the classes absent from SyntheticPizza10, to construct RealPizza10.
Similarly, MS-COCO-sub and VOC2007-sub datasets are constructed by removing images without having at least one class from the $\mathcal{S}$ domains. 
Please see \autoref{sec:add_benck} for details.  

\begin{figure}[t]
\centering
\includegraphics[trim={0cm +1cm 0cm +0cm},clip,scale=0.23]{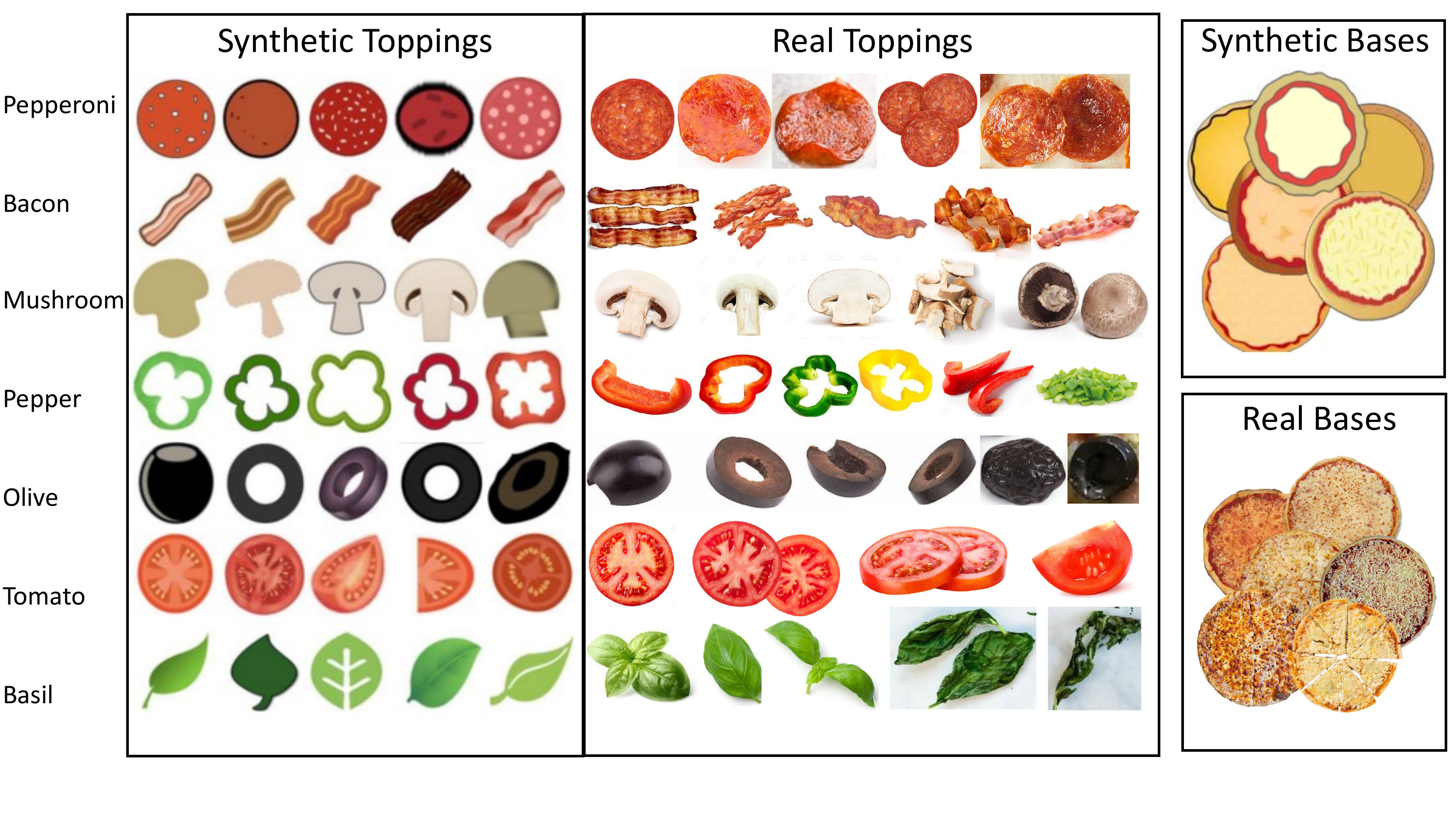}
\caption{\textbf{There is a domain shift in toppings and bases of pizzas.} Left: Examples of toppings used to create synthetic pizza images~\cite{papadopoulos2019make}. Middle: Examples of toppings in real pizza images. Right top: Examples of bases used to create synthetic pizza images~\cite{papadopoulos2019make}. Right bottom: Examples of bases in real pizza images.
}
\label{fig:toppings}
\end{figure}

The number of instances for each class in each pair of datasets
is unbalanced. Compared with the other benchmarks, SyntheticPizza10 $\rightarrow$ RealPizza10 are more challenging since all Pizza object instances are quite small and have diverse shape and texture. Although~\cite{papadopoulos2019make} uses a variety of different clip-art images for each topping to obtain the synthetic pizzas as shown in~\autoref{fig:toppings}, the number of these ingredient templates is still limited. In a real food image, the shape, color and texture of each ingredient object are dependent on cooking actions. As shown in~\autoref{fig:toppings}, for each ingredient, the domain gap between SyntheticPizza10 and RealPizza10 varies. 
In addition, the gap extends to bases of synthetic and real pizzas, as shown in~\autoref{fig:toppings}.

\noindent\textbf{Baselines and Evaluation Procedure.} 
We mainly focus on comparing against the state-of-the-art 
DAOD baselines (cross-domain):  \textbf{DT+PL}~\cite{inoue2018cross} and \textbf{PADOD}~\cite{hsu2020progressive}, and widely-used 
WSOD baselines (single-domain): \textbf{OICR}~\cite{tang2017multiple}, \textbf{CASD}~\cite{huang2020comprehensive}, and other baselines including WSDDN~\cite{bilen2016weakly}, PCL~\cite{tang2018pcl}, C-MIL~\cite{wan2019c}, WSOD2(+Reg)~\cite{zeng2019wsod2}, Pred Net~\cite{arun2019dissimilarity}, C-MIDN~\cite{gao2019c}, MIST(+Reg)~\cite{ren2020instance}, WeakRPN~\cite{tang2018weakly}, CASD$^2$ (training CASD two times: once for proposal and once for object detection), and CASD+W2N~\cite{huang2022w2n}.
Our evaluation follows the standard detection procedure. We compute Average Precision (\textbf{AP}) and the mean of AP (\textbf{mAP}) as the evaluation metric. A predicted box is treated as a positive example if it has an IOU $>$ $0.5$ between ground truth bounding boxes and the predicted box.

\noindent\textbf{Implementation Details.} 
In the warm-up stage, Faster R-CNN~\cite{ren2015faster} and Sparse DETR~\cite{roh2021sparse} 
were used as our FSOD models. 
For each target image, we generated 438 object proposals per image on average on VOC2007 and 441 object proposals per image on average on RealPizza10. Please refer to \autoref{sec:add_imple} for details.

\noindent\textbf{Source vs.\ Target Labeling Cost.} Two factors determine the trade-off of source vs.\ target FSOD: the cost of building a synthetic image generator and the realism of the synthesized images. When the realism of the synthetic images is moderate,  the cost of building the generator is low.  For SyntheticPizza10, built from abstract clipart or patches, the cost of generation is low. Moreover, the annotation of synthetic images is either a byproduct of the generation or inherently easy for human annotators, if such annotation is needed (\eg, Clipart1k).  Thus, our approach has inherently lower cost than the direct annotating the target domain.

\begin{table*}[hbt!]
\centering
\caption{Results (mAP in $\%$) for different methods on Clipart1K $\rightarrow$ VOC2007. We denote as Upper-Bound the FSOD (Faster R-CNN or Sparse DETR) results, trained and tested on {\em fully-annotated} target domain to indicate the weak upper-bound performance of our methods. Our warm-up stage is compared with CD models and our main stage is compared with SD models.
Faster R-CNN in CD means that we trained our network on fully-annotated source and test on fully-annotated target domains.
The best and second best results for {\DFOSDA} compared with baselines are shown in red and blue.
}
\vspace{-3mm}
\label{tab:main}

\centering
\label{tab:clip_voc_ab}
\footnotesize
\resizebox{2.1\columnwidth}{!}{
\begin{tabular}{c|c|c|c|c|c||c|c|c|c|c|c|c|c|c|c|c|c|c|c|c}
&  & \multicolumn{4}{c||}{After Warm-Up Stage} & \multicolumn{14}{c}{After Main Stage}\\
\hline
Type & & \multicolumn{3}{c|}{CD} & Ours & \multicolumn{12}{c|}{SD} & \multicolumn{2}{c}{Ours} \\
\hline
Method & Upper-Bound & Faster R-CNN~\cite{ren2015faster} & DT+PL~\cite{inoue2018cross} & PADOD~\cite{hsu2020progressive}  & ${\DFOSDA}_\text{warm-up}$  & WSDDN~\cite{bilen2016weakly} & OICR~\cite{tang2017multiple} & PCL~\cite{tang2018pcl} & WeakRPN~\cite{tang2018weakly}& C-MIL~\cite{wan2019c}
& WSOD2(+Reg)~\cite{zeng2019wsod2} & Pred Net~\cite{arun2019dissimilarity} & C-MIDN~\cite{gao2019c} & MIST(+Reg)~\cite{ren2020instance} 
& CASD~\cite{huang2020comprehensive} & CASD$^2$ & CASD+W2N~\cite{huang2022w2n}& ${\DFOSDA}_\text{casd}$ & ${\DFOSDA}_\text{casd+w2n}$\\
\hline
mAP & \em 69.9 & 22.8 & 34.6 & 24.2 & 37.3 & 34.8 & 41.2 & 43.5 & 45.3 & 50.5 & 53.6 & 52.9 & 52.6  & 54.9 & 57.0 & 57.4 & \color[rgb]{0,0,1} 65.4 & 64.8 & \color[rgb]{1,0,0} 66.9\\
\end{tabular}
}
\end{table*}

\begin{table*}[hbt!]
\centering
\caption{ Results (mAP in $\%$) for different methods on SyntheticPizza10 $\rightarrow$ RealPizza10 (SPizza $\rightarrow$ RPizza), Watercolor2K $\rightarrow$ VOC2007-sub (Water $\rightarrow$ VocS), Comic2K $\rightarrow$ VOC2007-sub (Comi $\rightarrow$ VocS), and Clipart1K $\rightarrow$ MS-COCO-sub (Clip $\rightarrow$ CocoS).}
\vspace{-3mm}
\label{tab:pizza_ab_other}
\footnotesize
\resizebox{1.9\columnwidth}{!}{
\footnotesize
\begin{tabular}{c|c|c|c|c|c|c||c |c|c }
&  & & \multicolumn{4}{c||}{After Warm-Up Stage} & \multicolumn{3}{c}{After Main Stage}\\
\hline
& Type & & \multicolumn{3}{c|}{CD} & Ours & \multicolumn{2}{c|}{SD} & Ours \\
\hline
& Method & Upper-Bound & Faster R-CNN~\cite{ren2015faster} & DT+PL~\cite{inoue2018cross} & PADOD~\cite{hsu2020progressive} & ${\DFOSDA}_\text{warm-up}$ & OICR~\cite{tang2017multiple} & CASD~\cite{huang2020comprehensive} & ${\DFOSDA}_\text{casd}$ \\
\hline
\multirow{4}{*}{mAP} & SPizza $\rightarrow$ RPizza & - & 4.3 & 14.9 & 8.1 & \color[rgb]{0,0,1}17.9 & 4.7 &  12.9 & \color[rgb]{1,0,0}25.1 \\
\cline{2-10}
& Water $\rightarrow$ VocS & \em 78.0 & 42.1 & 49.4 & - & 52.1 & - & \color[rgb]{0,0,1} 65.2 & \color[rgb]{1,0,0}73.2 \\
\cline{2-10}
& Comi $\rightarrow$ VocS & \em 78.0 & 33.5 & 46.5 & - & 49.6 & - &  \color[rgb]{0,0,1}65.2 & \color[rgb]{1,0,0}70.8 \\
\cline{2-10}
& Clip $\rightarrow$ CocoS & \em 84.3 & 13.9 & 22.1 & - & 25.7 & - & \color[rgb]{0,0,1}48.3 & \color[rgb]{1,0,0}57.2 \\
\end{tabular}
}
\vspace{-3mm}
\end{table*}

\begin{figure}[t]
    \centering
    \begin{minipage}{\linewidth}
  \begin{minipage}{0.32 \linewidth}
    \centering
    \includegraphics[trim={0 53cm 5cm 0cm},clip,width = .7\linewidth]{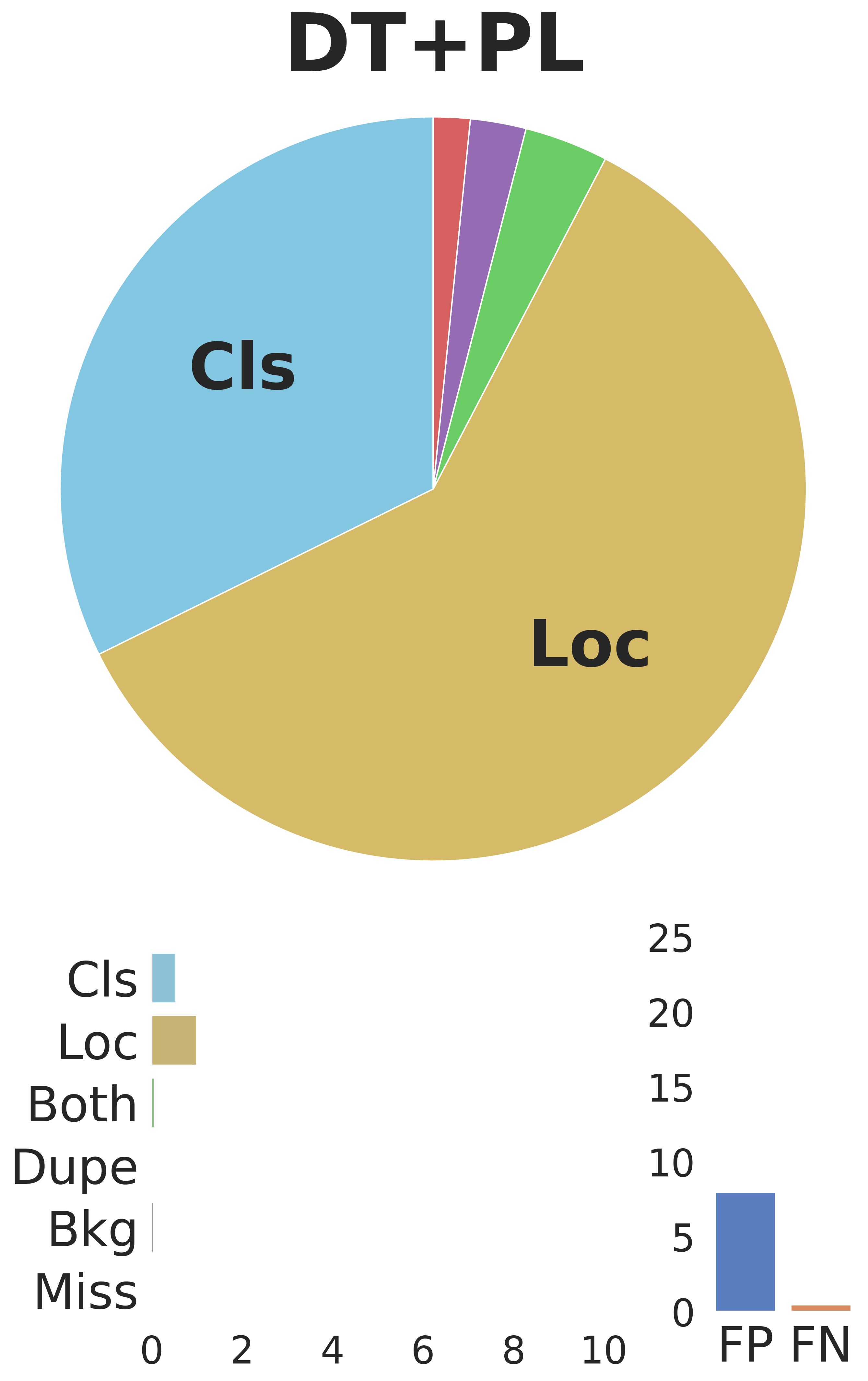}
    \end{minipage}
  \begin{minipage}{0.32 \linewidth}
    \centering
    \includegraphics[trim={0 53cm 5cm 0cm},clip,width = .7\linewidth]{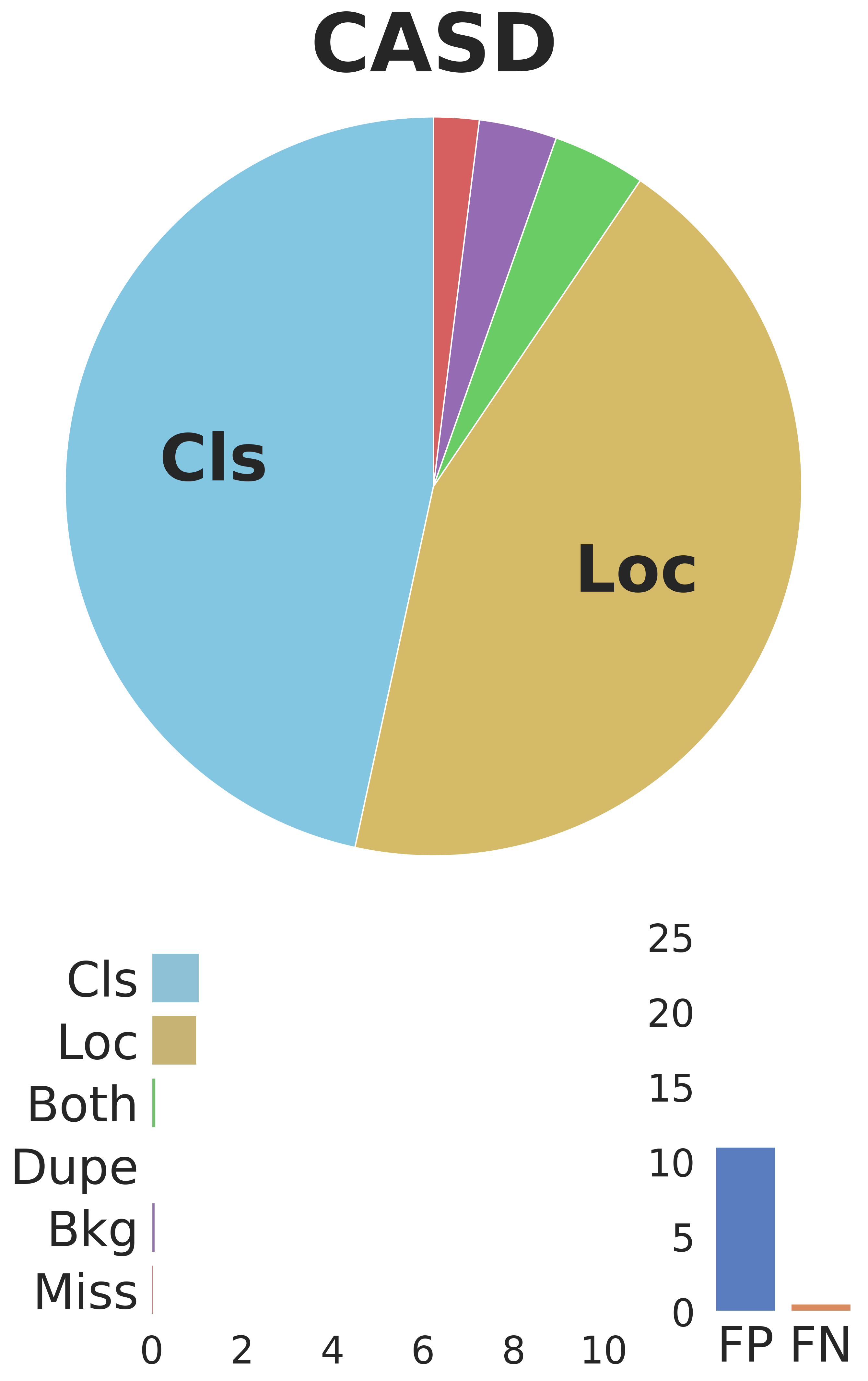}
  \end{minipage}
  \begin{minipage}{0.32 \linewidth}
    \centering
    \includegraphics[trim={0 53cm 5cm 0cm},clip,width = .7\linewidth]{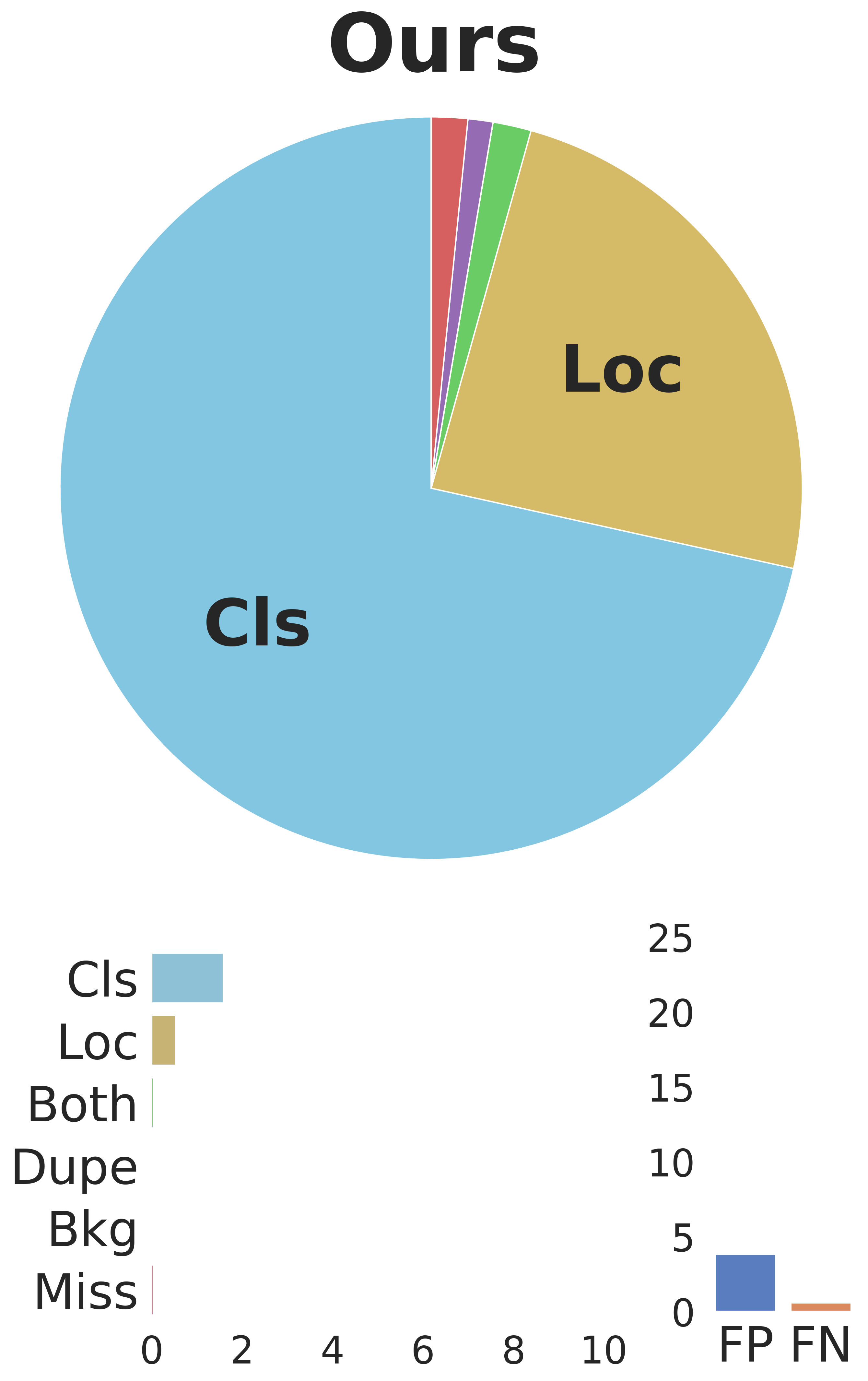}
  \end{minipage}
  \hfill
\end{minipage}
    \caption{Identify object detection errors.}
    \label{fig:error}
\end{figure}

\subsection{Main Results}
\label{sec:main_results}
We compare 
{\DFOSDA} with state-of-the-art
single (SD) and cross-domain (CD) methods in terms of mAP. \autoref{tab:clip_voc_ab} and \autoref{tab:pizza_ab_other} summarize the detection results on five benchmarks
based on Faster R-CNN FSOD backbone. The per class APs are listed in \autoref{tab:voc_pizza_baseline}. {\DFOSDA} incorporated with OICR is denoted as ${\DFOSDA}_\text{oicr}$, and with CASD is denoted as ${\DFOSDA}_\text{casd}$. The results of our warm-up stage are denoted as ${\DFOSDA}_\text{warm-up}$. 

\noindent\textbf{{\DFOSDA} consistently outperforms the SD baselines.} 
As shown in~\autoref{tab:clip_voc_ab}, on Clipart1K $\rightarrow$ VOC2007, ${\DFOSDA}_\text{casd}$ reaches $64.8\%$ mAP, outperforming the original CASD by $7.8\%$ mAP, and ${\DFOSDA}_\text{casd+w2n}$ reaches $66.9\%$ mAP, outperforming the original CASD+W2N by $1.5\%$ mAP, while CASD$^2$ outperforms the original CASD only by $0.4\%$ mAP. The detection performance does not benefit much from using CASD$^2$, since doing so does not improve the generated proposals.
On SyntheticPizza10 $\rightarrow$ RealPizza10 reported in~\autoref{tab:pizza_ab_other}, ${\DFOSDA}_\text{casd}$ provides a $12.2\%$ improvement over the original CASD in terms of mAP.
\textbf{{${\DFOSDA}$} also consistently outperforms the CD baselines.} \autoref{tab:clip_voc_ab} shows that on Clipart1K $\rightarrow$ VOC2007
${\DFOSDA}_\text{casd}$ outperforms DT+PL and PADOD by $30.2\%$ and $40.6\%$ mAP, respectively. 
As shown in \autoref{tab:pizza_ab_other}, ${\DFOSDA}_\text{casd}$ outperforms DT+PL and PADOD on SyntheticPizza10 $\rightarrow$ RealPizza10 by $10.2\%$ and $17.0\%$ mAP, respectively. \textbf{\noindent\textbf{{\DFOSDA} generalizes across different datasets.}} As shown in \autoref{tab:pizza_ab_other}, ${\DFOSDA}$ effectively handles different domain shifts, successfully leveraging a variety of $\mathcal{S}$\footnote{Resource constraints limit our focus on best select SOTA, with extensive comparison delegated to Clipart1K $\rightarrow$ VOC2007 evaluation.}.

We observe both stages of {\DFOSDA} yield consistently improved detection and localization performance compared with both state-of-the-art SD and CD baselines, especially on the more challenging SyntheticPizza10 $\rightarrow$ RealPizza10 scenario. By exploiting our domain adaptation stage, we believe that our training of the WSOD model is superior to existing methods in three important ways. First, our pre-trained features are target-domain-specific, because of progressive adaptation from source to intermediate to target domains, whereas existing WSOD methods use features pre-trained on ImageNet. Second, our pre-trained features are detection-aware, while ImageNet features used in existing WSOD methods are pre-trained with a single whole-image classification loss, which encourages translation and scale-invariant features. In contrast, the training of our FSOD model involves classification and regression losses, providing features that are sensitive to object locations and scales and are thus useful for detection. Third, our object proposals are target-domain-specific and of high-quality, since they are progressively learned directly on the target domain from foreground and background. 
Existing WSOD methods use hand-crafted selective search 
object proposals, which leads to inaccurate proposals especially for domains such as Pizza, with properties different from VOC2007.

\subsection{Ablation Study}
\label{sec:ablation_study}
We first conducted ablation studies to investigate the effectiveness of our warm-up stage on SyntheticPizza10 $\rightarrow$ RealPizza10 based on the Faster R-CNN FSOD backbone.

\noindent\textbf{Effectiveness of Progressive Adaptation.} In our warm-up stage, each adaptation stage (from FSOD-2 to FSOD-5) provides an improvement of 5.4, 0.6, 4.7, 2.9$\%$ compared with the previous step in terms of mAP, respectively. Therefore, each adaptation step in our warm-up stage is helpful. 

\noindent\textbf{Impact of Adaptation Order.} It is important {\em when} to use copy-paste augmentation. Starting from the same baseline model FSOD-1, if we sequentially fine-tune the FSOD-1 model on intermediate domain $\mathcal{G}_1$ and augmented intermediate domain $\mathcal{G}_2$, the detection performance will be improved by 6.0$\%$ mAP from FSOD-1 to FSOD-3. However, if we sequentially fine-tune the FSOD-1 model on augmented intermediate domain $\mathcal{G}_2$ and intermediate domain $\mathcal{G}_1$, the detection performance will be improved by only 1.6$\%$ mAP from FSOD-1 to FSOD-3. Similarly, starting from the same FSOD-3 model, if we sequentially fine-tune the FSOD-3 model on first-round pseudo-labeled domain $\mathcal{T}$ and second-round augmented pseudo-labeled domain $\mathcal{T}$, the detection performance will be improved 7.6$\%$ mAP from FSOD-3 to FSOD-5. However, if we sequentially fine-tune the FSOD-3 model on augmented first-round pseudo-labeled domain $\mathcal{T}$ and second-round pseudo-labeled domain $\mathcal{T}$, the detection performance will be improved by 7.0$\%$ mAP from FSOD-3 to FSOD-5.

\noindent\textbf{Generalizability of the Warm-up Stage across FSODs.} We investigate our warm-up stage on other FSOD models such as Sparse DETR on SyntheticPizza10 $\rightarrow$ RealPizza10 datasets. Compared with Faster R-CNN backbone, our ${\DFOSDA}_\text{warm-up}$ and ${\DFOSDA}_\text{casd}$ based on Sparse DETR yields 0.6$\%$ and 1.1$\%$ improvement in terms of mAP, respectively. These results emphasize the generality of our framework across different FSOD models. 
We also conduct ablation studies to investigate the effectiveness of our architecture components in the main FSOD stage, including the domain specific pre-trained deep FE and the weakly-supervised OP generator, as well as the generalization ability of our framework on two WSODs: OICR and CASD. We perform experiments on Clipart1K $\rightarrow$ VOC2007 and SyntheticPizza10 $\rightarrow$ RealPizza10. We find that: (1) our domain specific pre-trained deep FE and weakly-supervised OP generator are both necessary for \DFOSDA; and (2) \DFOSDA can generalize to different WSOD methods. 

\noindent\textbf{Main Stage Configurations.} From \autoref{tab:ablation}, we observe that compared with the single-domain baseline networks (OICR and CASD), replacing the VGG16 backbone pre-trained on ImageNet with \textbf{domain specific pre-trained deep FE}  can improve the performance on VOC2007 (mAP from $41.2\%$ to $44.7\%$, and from $57.0\%$ to $60.0\%$, respectively), and on RealPizza10 a consistent improvement is achieved with $3.8\%$ and $1.9\%$ for OICR and CASD,
respectively. From \autoref{tab:clip_voc_ab}, we observe that our object proposal generator is also better than WeakRPN~\cite{tang2018weakly}, including a two-stage region
proposal network. These results confirm the necessity of the domain-specific pre-trained deep features. 
\autoref{tab:ablation} also shows the impact of the \textbf{weakly-supervised OP generator}; it achieves consistent improvements of 3.1\% and 11.1\%, compared with CASD, on VOC2007 and RealPizza10 datasets, respectively. Together, $\textbf{FE+OP}$ results in \autoref{tab:ablation} suggest that these two key components are both effective and complementary to each other.

\begin{table}[t]
\centering
\caption{Ablation study of {\DFOSDA} main configurations on Clipart1K $\rightarrow$ VOC2007 and SyntheticPizza10 $\rightarrow$ RealPizza10.}\label{tab:ablation}
\vspace{-3mm}
    \begin{subtable}{\linewidth}
    \centering
    \footnotesize
    \resizebox{.9\columnwidth}{!}{
    \begin{tabular}{c|c|c|c}
     & & \multicolumn{2}{c}{mAP} \\
     \hline
    Type & Method & Clip $\rightarrow$ Voc & SPizza $\rightarrow$ RPizza\\
    \hline
    \hline
    SD & OICR & 41.2 & 4.7
    \\
    \hline
    \multirow{3}{*}{\texttt{${\DFOSDA}_\text{oicr}$}} & \textbf{+FE} & 44.7 & 8.5
    \\
    \cline{2-4}
    & \textbf{+OP} & 47.2 & 12.6
    \\
    \cline{2-4}
    & \textbf{+FE+OP} & 52.7 & 13.8
    \\
    \hline
    \hline
    SD & CASD & 57.0 & 12.9\\
    \hline
    \multirow{3}{*}{\texttt{${\DFOSDA}_\text{casd}$}} & \textbf{+FE} & 60.0 & 14.8\\
    \cline{2-4}
    & \textbf{+OP} & 60.1 & 24.0\\  
    \cline{2-4}
    & \textbf{+FE+OP} & \color[rgb]{1,0,0}64.8 & \color[rgb]{1,0,0}25.1
    \\
    \end{tabular}
    }
    \end{subtable}
~
\vspace{-3mm}
\end{table}

\begin{figure}[htp]
  \centering
  \includegraphics[trim={0cm 9cm 0 0},clip,scale=0.24]{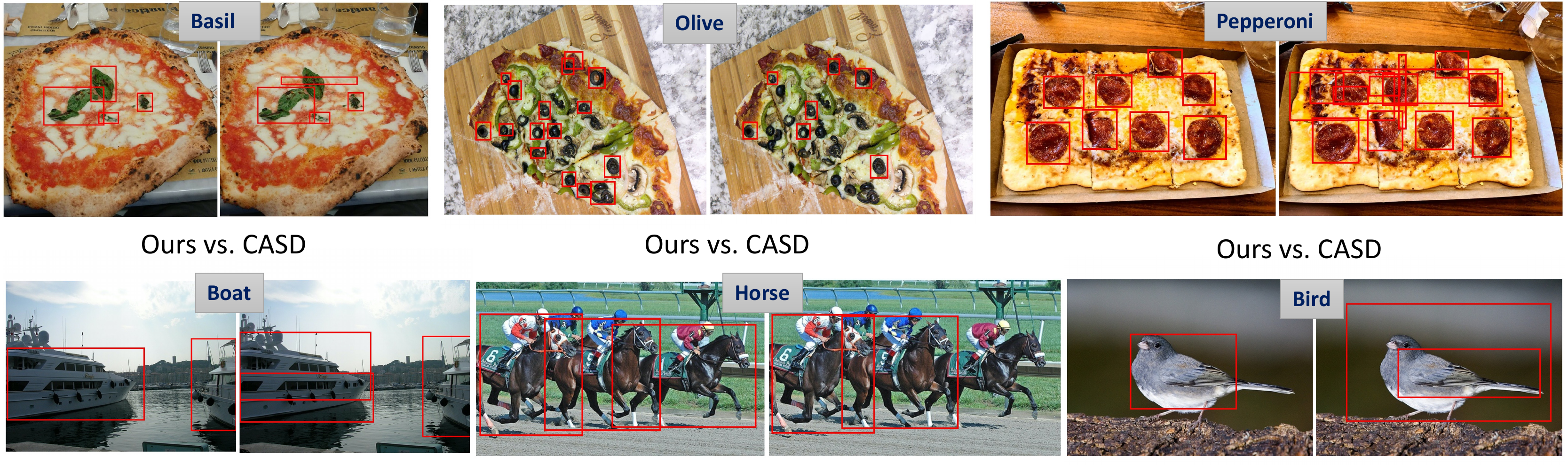}
  \vspace{-3mm}
   \caption{Example of success cases for our ${\DFOSDA}_\text{casd}$ vs. CASD in the test set of RealPizza10 and VOC2007 datasets.
   We only show instances with scores over 0.3 to maintain visibility.}
   \label{fig:positives}
   \vspace{-3mm}
\end{figure}

\noindent\textbf{Generalizability of \DFOSDA across WSODs.} 
We investigate
the impact of our framework as a function of different WSOD methods (here, OICR and CASD). Results in \autoref{tab:ablation} emphasize the \textbf{generalizability of \DFOSDA across WSODs}. The performance gain is observed in both OICR and CASD on the two datasets. The effect of \DFOSDA is particularly significant for CASD on RealPizza 10, since our object proposals are target domain-specific and of high-quality. By contrast, existing WSOD methods use hand-crafted selective search to generate object proposals, leading to inaccurate proposals especially for domains such as Pizza that are very dissimilar to VOC2007.

\noindent\textbf{Identifying Object Detection
Errors.} We use TIDE~\cite{bolya2020tide} to understand the {\color{blue}{classification}}, {\color{brown}{localization}}, {\color{green}{both Cls and Loc}}, {\color{pink}{duplicate detection}}, {\color{violet}{background}}, and {\color{red}{missed GT}} errors in our model. As shown in~\autoref{fig:error}, \DFOSDA effectively reduces the localization error. Please see \autoref{sec:add_ablation} for more details.

\subsection{Qualitative Analysis}
\autoref{fig:positives} illustrates the detection results produced by our \DFOSDA and CASD on RealPizza10 and VOC2007 datasets, respectively. 
There, it can be observed that \DFOSDA does not only locate most objects, but that it also produces more accurate bounding boxes. Specifically, in the RealPizza10 images it can be appreciated bounding boxes provided by our method (left) closely align with the objects of interest, while for CASD (right) bounding boxes are often imprecise (either wrong shape or big/small). Similar observations can be made for VOC2007 where CASD often fails to locate objects or produces spurious bounding boxes.

\section{Discussion and Conclusion}
We propose \DFOSDA, a simple yet effective object generation strategy that can be applied to different WSOD methods. The key insight is to cast WSOD as a domain adaptation problem and improve performance by progressive foreground-background focused transfer learning of an FSOD from non-photographic source to real-world target domains.
Empirical evaluation shows \DFOSDA significantly outperforms state of the art on several benchmarks.  

\noindent\textbf{Limitation.} Our framework requires extra training time for CycleGAN, which brings in the most additional computation overhead.
While \DFOSDA  offers a promising way to solve WSOD in the presence of a large domain gap, it currently lacks the ability to  jointly learn and refine all stages in the pipeline. An end-to-end large-gap WSOD could offer additional improvement in detection performance on the target domain through creation of increasingly discriminative object features.  However, one challenge 
with that setting
would be to control the back-propagation of possible errors induced by the PL steps.

\noindent\textbf{Acknowledgement}. This work was supported in part by NSF IIS Grant \#1955404.

{\small
\bibliographystyle{ieee_fullname}
\bibliography{egbib}

\begin{thebibliography}{10}\itemsep=-1pt

\bibitem{arun2019dissimilarity}
Aditya Arun, CV Jawahar, and M~Pawan Kumar.
\newblock Dissimilarity coefficient based weakly supervised object detection.
\newblock In {\em CVPR}, 2019.

\bibitem{bilen2016weakly}
Hakan Bilen and Andrea Vedaldi.
\newblock Weakly supervised deep detection networks.
\newblock In {\em CVPR}, 2016.

\bibitem{bolya2020tide}
Daniel Bolya, Sean Foley, James Hays, and Judy Hoffman.
\newblock Tide: A general toolbox for identifying object detection errors.
\newblock In {\em ECCV}, 2020.

\bibitem{carion2020end}
Nicolas Carion, Francisco Massa, Gabriel Synnaeve, Nicolas Usunier, Alexander
  Kirillov, and Sergey Zagoruyko.
\newblock End-to-end object detection with transformers.
\newblock In {\em ECCV}, 2020.

\bibitem{chen2018domain}
Yuhua Chen, Wen Li, Christos Sakaridis, Dengxin Dai, and Luc Van~Gool.
\newblock Domain adaptive faster r-cnn for object detection in the wild.
\newblock In {\em CVPR}, 2018.

\bibitem{dietterich1997solving}
Thomas~G Dietterich, Richard~H Lathrop, and Tom{\'a}s Lozano-P{\'e}rez.
\newblock Solving the multiple instance problem with axis-parallel rectangles.
\newblock {\em Artificial intelligence}, 89(1-2):31--71, 1997.

\bibitem{everingham2010pascal}
Mark Everingham, Luc Van~Gool, Christopher~KI Williams, John Winn, and Andrew
  Zisserman.
\newblock The pascal visual object classes (voc) challenge.
\newblock {\em IJCV}, 88(2):303--338, 2010.

\bibitem{gao2019c}
Yan Gao, Boxiao Liu, Nan Guo, Xiaochun Ye, Fang Wan, Haihang You, and Dongrui
  Fan.
\newblock C-midn: Coupled multiple instance detection network with segmentation
  guidance for weakly supervised object detection.
\newblock In {\em ICCV}, 2019.

\bibitem{ghiasi2021simple}
Golnaz Ghiasi, Yin Cui, Aravind Srinivas, Rui Qian, Tsung-Yi Lin, Ekin~D Cubuk,
  Quoc~V Le, and Barret Zoph.
\newblock Simple copy-paste is a strong data augmentation method for instance
  segmentation.
\newblock In {\em CVPR}, 2021.

\bibitem{girshick2015fast}
Ross Girshick.
\newblock Fast r-cnn.
\newblock In {\em ICCV}, 2015.

\bibitem{he2016deep}
Kaiming He, Xiangyu Zhang, Shaoqing Ren, and Jian Sun.
\newblock Deep residual learning for image recognition.
\newblock In {\em CVPR}, 2016.

\bibitem{hsu2020progressive}
Han-Kai Hsu, Chun-Han Yao, Yi-Hsuan Tsai, Wei-Chih Hung, Hung-Yu Tseng, Maneesh
  Singh, and Ming-Hsuan Yang.
\newblock Progressive domain adaptation for object detection.
\newblock In {\em WACV}, 2020.

\bibitem{huang2022w2n}
Zitong Huang, Yiping Bao, Bowen Dong, Erjin Zhou, and Wangmeng Zuo.
\newblock W2n: Switching from weak supervision to noisy supervision for object
  detection.
\newblock In {\em ECCV}, 2022.

\bibitem{huang2020comprehensive}
Zeyi Huang, Yang Zou, BVK Kumar, and Dong Huang.
\newblock Comprehensive attention self-distillation for weakly-supervised
  object detection.
\newblock In {\em NeurIPS}, 2020.

\bibitem{inoue2018cross}
Naoto Inoue, Ryosuke Furuta, Toshihiko Yamasaki, and Kiyoharu Aizawa.
\newblock Cross-domain weakly-supervised object detection through progressive
  domain adaptation.
\newblock In {\em CVPR}, 2018.

\bibitem{kim2019self}
Seunghyeon Kim, Jaehoon Choi, Taekyung Kim, and Changick Kim.
\newblock Self-training and adversarial background regularization for
  unsupervised domain adaptive one-stage object detection.
\newblock In {\em ICCV}, 2019.

\bibitem{lee2013pseudo}
Dong-Hyun Lee et~al.
\newblock Pseudo-label: The simple and efficient semi-supervised learning
  method for deep neural networks.
\newblock In {\em ICML Workshop}, 2013.

\bibitem{lin2014microsoft}
Tsung-Yi Lin, Michael Maire, Serge Belongie, James Hays, Pietro Perona, Deva
  Ramanan, Piotr Doll{\'a}r, and C~Lawrence Zitnick.
\newblock Microsoft coco: Common objects in context.
\newblock In {\em ECCV}, 2014.

\bibitem{nguyen2009weakly}
Minh~Hoai Nguyen, Lorenzo Torresani, Fernando De~La~Torre, and Carsten Rother.
\newblock Weakly supervised discriminative localization and classification: a
  joint learning process.
\newblock In {\em ICCV}, 2009.

\bibitem{papadopoulos2019make}
Dim~P Papadopoulos, Youssef Tamaazousti, Ferda Ofli, Ingmar Weber, and Antonio
  Torralba.
\newblock How to make a pizza: Learning a compositional layer-based gan model.
\newblock In {\em CVPR}, 2019.

\bibitem{ren2015faster}
Shaoqing Ren, Kaiming He, Ross Girshick, and Jian Sun.
\newblock Faster r-cnn: Towards real-time object detection with region proposal
  networks.
\newblock In {\em NeurIPS}, 2015.

\bibitem{ren2020instance}
Zhongzheng Ren, Zhiding Yu, Xiaodong Yang, Ming-Yu Liu, Yong~Jae Lee,
  Alexander~G Schwing, and Jan Kautz.
\newblock Instance-aware, context-focused, and memory-efficient weakly
  supervised object detection.
\newblock In {\em CVPR}, 2020.

\bibitem{rodriguez2019domain}
Adrian~Lopez Rodriguez and Krystian Mikolajczyk.
\newblock Domain adaptation for object detection via style consistency.
\newblock {\em arXiv preprint arXiv:1911.10033}, 2019.

\bibitem{roh2021sparse}
Byungseok Roh, JaeWoong Shin, Wuhyun Shin, and Saehoon Kim.
\newblock Sparse detr: Efficient end-to-end object detection with learnable
  sparsity.
\newblock In {\em ICLR}, 2022.

\bibitem{roychowdhury2019automatic}
Aruni RoyChowdhury, Prithvijit Chakrabarty, Ashish Singh, SouYoung Jin, Huaizu
  Jiang, Liangliang Cao, and Erik Learned-Miller.
\newblock Automatic adaptation of object detectors to new domains using
  self-training.
\newblock In {\em CVPR}, 2019.

\bibitem{russakovsky2015imagenet}
Olga Russakovsky, Jia Deng, Hao Su, Jonathan Krause, Sanjeev Satheesh, Sean Ma,
  Zhiheng Huang, Andrej Karpathy, Aditya Khosla, Michael Bernstein, et~al.
\newblock Imagenet large scale visual recognition challenge.
\newblock {\em IJCV}, 115(3):211--252, 2015.

\bibitem{saito2019strong}
Kuniaki Saito, Yoshitaka Ushiku, Tatsuya Harada, and Kate Saenko.
\newblock Strong-weak distribution alignment for adaptive object detection.
\newblock In {\em CVPR}, 2019.

\bibitem{simonyan2014very}
Karen Simonyan and Andrew Zisserman.
\newblock Very deep convolutional networks for large-scale image recognition.
\newblock In {\em ICLP}, 2015.

\bibitem{sindagi2020prior}
Vishwanath~A Sindagi, Poojan Oza, Rajeev Yasarla, and Vishal~M Patel.
\newblock Prior-based domain adaptive object detection for hazy and rainy
  conditions.
\newblock In {\em ECCV}, 2020.

\bibitem{szegedy2015going}
Christian Szegedy, Wei Liu, Yangqing Jia, Pierre Sermanet, Scott Reed, Dragomir
  Anguelov, Dumitru Erhan, Vincent Vanhoucke, and Andrew Rabinovich.
\newblock Going deeper with convolutions.
\newblock In {\em CVPR}, 2015.

\bibitem{tang2018pcl}
Peng Tang, Xinggang Wang, Song Bai, Wei Shen, Xiang Bai, Wenyu Liu, and Alan
  Yuille.
\newblock Pcl: Proposal cluster learning for weakly supervised object
  detection.
\newblock {\em TPAMI}, 42(1):176--191, 2018.

\bibitem{tang2017multiple}
Peng Tang, Xinggang Wang, Xiang Bai, and Wenyu Liu.
\newblock Multiple instance detection network with online instance classifier
  refinement.
\newblock In {\em CVPR}, 2017.

\bibitem{tang2018weakly}
Peng Tang, Xinggang Wang, Angtian Wang, Yongluan Yan, Wenyu Liu, Junzhou Huang,
  and Alan Yuille.
\newblock Weakly supervised region proposal network and object detection.
\newblock In {\em ECCV}, 2018.

\bibitem{uijlings2013selective}
Jasper~RR Uijlings, Koen~EA Van De~Sande, Theo Gevers, and Arnold~WM Smeulders.
\newblock Selective search for object recognition.
\newblock {\em IJCV}, 104(2):154--171, 2013.

\bibitem{van2017learning}
Nanne Van~Noord and Eric Postma.
\newblock Learning scale-variant and scale-invariant features for deep image
  classification.
\newblock {\em Pattern Recognition}, 61:583--592, 2017.

\bibitem{vs2021mega}
Vibashan VS, Vikram Gupta, Poojan Oza, Vishwanath~A Sindagi, and Vishal~M
  Patel.
\newblock Mega-cda: Memory guided attention for category-aware unsupervised
  domain adaptive object detection.
\newblock In {\em CVPR}, 2021.

\bibitem{wan2019c}
Fang Wan, Chang Liu, Wei Ke, Xiangyang Ji, Jianbin Jiao, and Qixiang Ye.
\newblock C-mil: Continuation multiple instance learning for weakly supervised
  object detection.
\newblock In {\em CVPR}, 2019.

\bibitem{yun2019cutmix}
Sangdoo Yun, Dongyoon Han, Seong~Joon Oh, Sanghyuk Chun, Junsuk Choe, and
  Youngjoon Yoo.
\newblock Cutmix: Regularization strategy to train strong classifiers with
  localizable features.
\newblock In {\em ICCV}, 2019.

\bibitem{zeng2019wsod2}
Zhaoyang Zeng, Bei Liu, Jianlong Fu, Hongyang Chao, and Lei Zhang.
\newblock Wsod2: Learning bottom-up and top-down objectness distillation for
  weakly-supervised object detection.
\newblock In {\em ICCV}, 2019.

\bibitem{zhang2021hierarchical}
Ming Zhang, Shuaicheng Liu, and Bing Zeng.
\newblock Hierarchical region proposal refinement network for weakly supervised
  object detection.
\newblock In {\em ICIP}, 2021.

\bibitem{zhang2019bag}
Zhi Zhang, Tong He, Hang Zhang, Zhongyue Zhang, Junyuan Xie, and Mu Li.
\newblock Bag of freebies for training object detection neural networks.
\newblock {\em arXiv preprint arXiv:1902.04103}, 2019.

\bibitem{zhao2020collaborative}
Ganlong Zhao, Guanbin Li, Ruijia Xu, and Liang Lin.
\newblock Collaborative training between region proposal localization and
  classification for domain adaptive object detection.
\newblock In {\em ECCV}, 2020.

\bibitem{zhu2017unpaired}
Jun-Yan Zhu, Taesung Park, Phillip Isola, and Alexei~A Efros.
\newblock Unpaired image-to-image translation using cycle-consistent
  adversarial networks.
\newblock In {\em ICCV}, 2017.

\bibitem{zitnick2014edge}
C~Lawrence Zitnick and Piotr Doll{\'a}r.
\newblock Edge boxes: Locating object proposals from edges.
\newblock In {\em ECCV}, 2014.

\end{thebibliography}
}

\clearpage
\appendix
\section{Details of WSOD Models}
\label{sec:wsod_detail}
Our object proposal generation strategy can be applied to different WSOD methods to boost their detection performance. In the main paper, we show two representative WSOD models -- the widely-used OICR~\cite{tang2017multiple} and the state-of-the-art CASD~\cite{huang2020comprehensive}. Here, we highlight the main modeling elements of OICR and CASD. Additional details can be found in~\cite{tang2017multiple,huang2020comprehensive}.

As mentioned in \autoref{sec:main_wsod} of the main paper, we obtain $d$-dimensional object proposal feature vectors $\mathbf{V}_t\in{\mathbb{R}}^{{d}{\times}{M_t}}$ for each input image $\mathbf{X}_t$, where $M_t$ is the number of the proposal bounding boxes associated with $\mathbf{X}_t$. These object features are fed into the detection head of OICR~\cite{tang2017multiple} or CASD~\cite{huang2020comprehensive} to classify and localize objects. Both models contain a core multiple instance detection network (MIL) and $P$ instance refinement classifiers.

\begin{figure*}[hbt!]
  \centering
  \includegraphics[trim={0cm +2cm 0cm 0cm},clip,scale=0.5]{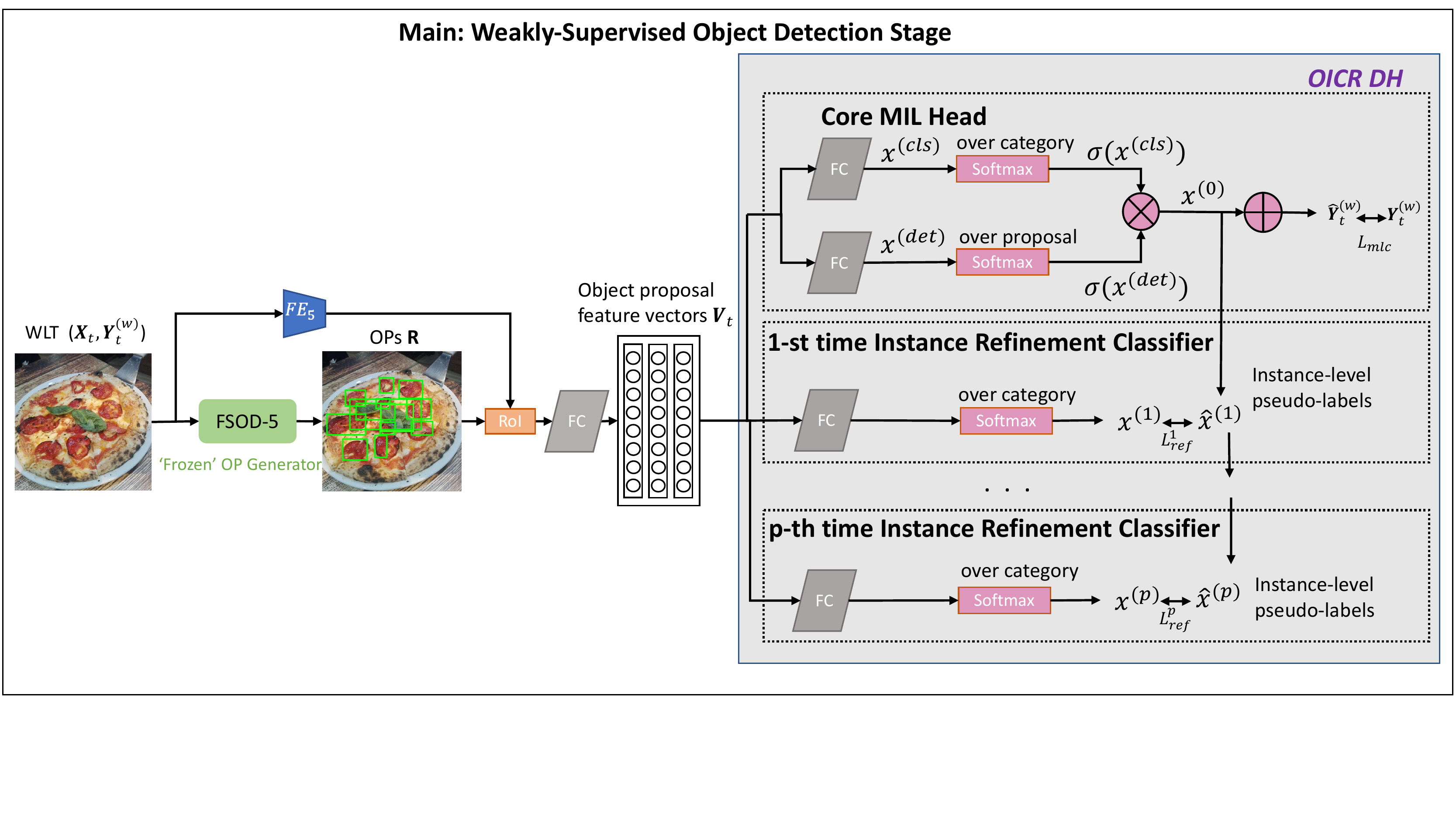}
  \vspace{-3mm}
   \caption{The architecture of our main weakly-supervised object detection stage.}
   \label{fig:oicr_ar}
   \vspace{-3mm}
\end{figure*}

\subsection{Core MIL Head}
As shown in~\autoref{fig:oicr_ar}, the core multiple instance detection network conducts the image-level multiple instance classification supervised by image-level labels $\mathbf{Y}^{(w)}_t$. In the core multiple instance detection network, the object proposal feature vectors $\mathbf{V}_t$ of image $\mathbf{X}_t$ are branched into two parallel classification and detection streams to generate two matrices $\mathbf{x}^\mathrm{(cls)}$ and $\mathbf{x}^\mathrm{(det)} \in{\mathbb{R}}^{C{\times}{M_t}}$ by two FC layers, where $C$ is number of classes in $\mathcal{T}$. Then, $\mathbf{x}^\mathrm{(cls)}$ passes through a softmax layer along the category direction (column-wise), while $\mathbf{x}^\mathrm{(det)}$ passes through another softmax layer along the proposal direction (row-wise), leading to $\sigma(\mathbf{x}^\mathrm{(cls)})$ and $\sigma(\mathbf{x}^\mathrm{(det)})$, respectively. The instance-level classification score for the object proposals is computed as the element-wise product $\mathbf{x}^{(0)} = \sigma(\mathbf{x}^\mathrm{(cls)}) \odot \sigma(\mathbf{x}^\mathrm{(det)})$. Finally, the image-level classification score for class $c$ is obtained as $p_c = \sum_{i=1}^{M_t} \mathbf{x}^{(0)}_{c,i}$. We train the core instance classifier using a multi-class cross-entropy loss $\mathcal{L}_\mathrm{mlc}$. By using the instance-level classification scores $\mathbf{x}^{(0)}$, we select proposals as detected objects. However, the core MIL head focuses on most discriminative object instances. 

\subsection{OICR} To address this issue, OICR~\cite{tang2017multiple} introduces multi-stage instance refinement classifiers
to refine the core instance classifier.
As shown in~\autoref{fig:oicr_ar}, $\mathbf{V}_t$ 
is
fed into $P$ refinement instance classifiers. Each $p$-th refinement classifier comprises of an FC and a softmax layers along the category direction, and produces a proposal score matrix $\mathbf{x}^{(p)} \in{\mathbb{R}}^{{(C+1)}{\times}{M_t}}$, where the $(C+1)$-th category is the background class. We train the $p$-th refinement instance classifier via a log loss $\mathcal{L}_\mathrm{ref}^p$ supervised by instance-level pseudo-labels, which are selected from the top-scoring proposals in the previous stage.

The loss for training the OICR network $\mathcal{L}_\mathrm{oicr}$ is defined as
\begin{equation}
\mathcal{L}_\mathrm{oicr} = \mathcal{L}_\mathrm{mlc} +  \lambda_{d} \sum_{p=1}^{P} \mathcal{L}_\mathrm{ref}^p, 
\end{equation}
where $\lambda_{d}$ is the trade-off hyperparameter.

\subsection{CASD} To further improve OICR, CASD~\cite{huang2020comprehensive} employs an attention-based feature learning method for WSOD model training. 
In addition to the $\mathcal{L}_\mathrm{oicr}$ loss, CASD includes a proposal bounding box smooth $\mathcal{L}_1$ regression loss $\mathcal{L}_\mathrm{reg}^p$ for $p$-th refinement instance classifier by following~\cite{zeng2019wsod2,ren2020instance}. 

To encourage consistent representation learning of the same image under different transformations (horizontal flipping and scaling), we consider the Input-wise CASD following~\cite{huang2020comprehensive}. For each image $\mathbf{X}_t$, we construct a set of images $\mathbf{X}_t^{tr} = \{\mathbf{X}_t^{(s1)},\mathbf{X}_t^{(\mathrm{flip}(s1))}, \ldots, \mathbf{X}_t^{(sn)},\mathbf{X}_t^{(\mathrm{flip}(sn))}\}$, where $\mathbf{X}_t^{(si)}$ is its scaled image at $si$ scale, $\mathbf{X}_t^{(\mathrm{flip}(si))}$ is the horizontally flipped image of the scaled image, and $n$ is the number of scales. Then, by feeding the set of images $\mathbf{X}_t^{tr}$ into the same feature extractor $FE_5$ in the WSOD model, we obtain a set of image feature maps $\mathbf{F}_t^{tr} = \{\mathbf{F}_t^{(s1)},\mathbf{F}_t^{(\mathrm{flip}(s1))}, \ldots, \mathbf{F}_t^{(sn)},\mathbf{F}_t^{(\mathrm{flip}(sn))}\}$. For each object proposal $r$, we compute the proposal feature vectors cropped from $\mathbf{F}_t^{tr}$, and use all proposal feature vectors to obtain a set of object proposal attention maps $\mathbf{A}_r^{tr} = \{\mathbf{A}_r^{(s1)},\mathbf{A}_r^{(\mathrm{flip}(s1))}, \ldots, \mathbf{A}_r^{(sn)},\mathbf{A}_r^{(\mathrm{flip}(sn))}\}$ by channel-wise average pooling and element-wise sigmoid function. We use the aggregated attention maps $\mathbf{A}_r^{IW} = max(\mathbf{A}_r^{tr})$, where $max(\cdot)$ is the element-wise max operator, to update the feature extractor $FE_5$ in the $p$-th refinement step. 
During training, we add the $\mathcal{L}_\mathrm{IW}^p$ loss, which is an $\mathcal{L}_2$ loss between $\mathbf{A}_r^{IW}$ and each object proposal attention map in $\mathbf{A}_r^{tr}$.

To encourage balanced representation learning of the same image produced at different feature extractor layers, we consider the Layer-wise CASD following~\cite{huang2020comprehensive}. The feature extractor $FE_5$ consists of $Q$ number of convolutional blocks $FE_5 = \{B_1, \ldots, B_Q\}$. The original image $\mathbf{X}_t$ is fed into each of them to output a set of feature map $\mathbf{F}_t^{B} = \{\mathbf{F}_t^{B_1}, \ldots, \mathbf{F}_t^{B_Q}\}$. For each object proposal $r$, we compute the proposal feature vector in each block and use all the feature vectors to obtain a set of object proposal attention maps $\mathbf{A}_r^{bl} = \{\mathbf{A}_r^{B_1}, \ldots, \mathbf{A}_r^{B_Q}\}$. Similarly to Input-wise CASD, we obtain the aggregated attention maps $\mathbf{A}_r^{LW} = max(\mathbf{A}_r^{bl})$. To update the feature extractor $FE_5$ in the $p$-th refinement step, we add the $\mathcal{L}_\mathrm{LW}^p$ loss, which is an $\mathcal{L}_2$ loss between $\mathbf{A}_r^{LW}$ and each object proposal attention map in $\mathbf{A}_r^{bl}$.

The loss for training the CASD network $\mathcal{L}_\mathrm{casd}$ is defined as
\begin{equation}
\mathcal{L}_\mathrm{casd} = \mathcal{L}_\mathrm{mlc} +  \sum_{p=1}^{P} (\lambda_{d} \mathcal{L}_\mathrm{ref}^p + \lambda_{g} \mathcal{L}_\mathrm{reg}^p +  \lambda_{i} \mathcal{L}_\mathrm{IW}^p + \lambda_{i} \mathcal{L}_\mathrm{LW}^p),
\end{equation}
where $\lambda_{d}$, $\lambda_{g}$, and $\lambda_{i}$ are the trade-off hyperparameters. For additional details please refer to~\cite{tang2017multiple,huang2020comprehensive}.

\section{Benchmarks}
\label{sec:add_benck}
We evaluate our method on five dual-domain image benchmark pairs: SyntheticPizza10~\cite{papadopoulos2019make} $\rightarrow$ RealPizza10~\cite{papadopoulos2019make}, Clipart1K~\cite{inoue2018cross} $\rightarrow$ VOC2007~\cite{everingham2010pascal}, Watercolor2K~\cite{inoue2018cross} $\rightarrow$ VOC2007-sub, Comic2K~\cite{inoue2018cross} $\rightarrow$ VOC2007-sub, and Clipart1K $\rightarrow$ MS-COCO-sub~\cite{lin2014microsoft} datasets. 
Each synthetic pizza contains up to 10 toppings. The total 16,340 SyntheticPizza10 images are split into 14,802 training and 1,538 testing. RealPizza10 is split into 5,029 training and 552 testing images. Both Clipart1K and VOC2007 contain 20 object classes. Both Watercolor2K and Comic2K contain 6 classes: bike, bird, car, cat, dog, and person, the subset of classes in the VOC2007. The Watercolor2K and Comic2K domains are split into two subsets: 1,000 training and 1,000 testing images. The VOC2007-sub dataset includes 3,487 training and 3,457 testing images. The MS-COCO-sub domain includes 95,279 training images and 4,031 testing images. 

\noindent\textbf{Challenge of SyntheticPizza10 $\rightarrow$ RealPizza10 benchmark.} One of our contributions is that we construct the dual-domain benchmark SyntheticPizza10 $\rightarrow$ RealPizza10 for WSOD.  Compared with existing WSOD benchmarks (VOC2007 and MS-COCO), RealPizza10 is more challenging for the following reasons. First, there are more objects per image. On RealPizza10 each image contains 20 objects on average, while on VOC2007 there are 2 objects per image on average. Second, there is strong variation in appearance. For example, the toppings in Pizza images exhibit complex changes in appearance. Third, there are layered object occlusions due to the topping objects. Such difficulty is further reflected in the low detection performance of baselines on RealPizza10 (e.g., Faster R-CNN achieves only 4.3\% mAP on RealPizza10, while 22.8\% mAP on VOC2007 and 13.9\% mAP on MS-COCO-sub, as shown in Tables 1 and 2 in the main paper). We hope Pizza10 can serve as a new WSOD benchmark.

\section{Additional Implementation Details}
\label{sec:add_imple}
\paragraph{\textbf{\textup{Implementation Details.}}}
All models and experiments were implemented in Pytorch. The VGG16~\cite{simonyan2014very} and ResNet-50~\cite{he2016deep} models pre-trained on ImageNet~\cite{russakovsky2015imagenet} were used as FSOD feature extractor and WSOD feature extractor. VGG16 was used as Faster R-CNN feature extractor and ResNet-50 was used as Sparse DETR feature extractor. Because VOC2007 and MS-COCO lack clean background, we run experiments on Clipart1K $\rightarrow$ VOC2007, Watercolor2K $\rightarrow$ VOC2007-sub, Comic2K $\rightarrow$ VOC2007-sub, and Clipart1K $\rightarrow$ MS-COCO-sub without the 3rd adaptation step FSOD-3 on augmented intermediate domain $\mathcal{G}_{2}$ in our warm-up stage.

In the main stage, the maximum number of training iteration was set to be 100K for all target domains.
\paragraph{\textbf{\textup{Copy-Paste Augmentation.}}}
As stated in \autoref{sec:main_warmup} of the main paper, we employ an object-aware data augmentation method based on copy-paste~\cite{yun2019cutmix} to map images $\mathbf{X}_{g_1}$ to $\mathbf{X}_{g_2}$. For each image $\mathbf{X}_{g_1}$, we randomly copy several foreground object instances from $\mathcal{G}_1$, with resizing and flipping transformations, and paste them onto the real-world target background images from $\mathcal{T}$ to generate $\mathbf{X}_{g_2}$. The flipping transformation includes horizontal and vertical flipping transformations. The resize ratio is a random value between 0.8 to 1.2. The PL step can be in principle performed for $K$ times to generate instance-level pseudo-annotations. In our experiments, we found that running the PL step twice has achieved satisfactory performance.  After the second PL step, we apply the copy-paste augmentation with resizing and flipping transformations for the pseudo-labeled target images. According to the statistic information of Clipart1k $\rightarrow$ VOC2007 reported in \autoref{tab:num_instance_voc}, for each class, there is a maximum of two objects in an image. Therefore, we copy each pseduo-labeled object and randomly paste it 0 or 1 time onto the original image. On SyntheticPizza10 $\rightarrow$ RealPizza10, according to the statistic information reported in \autoref{tab:num_instance_pizza}, we copy each pseduo-labeled object and randomly paste it maximum 20 times onto the original image. All the pasted objects and original objects have no overlapping.

\paragraph{\textbf{\textup{CycleGAN.}}}
We trained CycleGAN~\cite{zhu2017unpaired} with the learning rate of $1.0 \times 10^{-5}$ for the first ten epochs and a linear decaying rate to zero over the next ten epochs following~\cite{inoue2018cross} to generate intermediate images. We followed the original paper~\cite{zhu2017unpaired} for remaining hyperparameters.

\paragraph{\textbf{\textup{Faster R-CNN.}}}
We trained Faster R-CNN~\cite{ren2015faster} on images of a single scale. The short edge of input images was re-scaled to 600, and the longest image edge was capped to $1000$. We employed a learning rate, which is the same as the final learning rate for the previous step, to progressively fine-tune Faster R-CNN on (1) a transfer-labeled intermediate domain $\mathcal{G}_1$, 
(2) augmented transfer-labeled intermediate domain $\mathcal{G}_2$ and 
then on (3) the pseudo-labeled target domain $\mathcal{T}$ and (4) augmented pseudo-labeled target domain $\mathcal{T}$.

\paragraph{\textbf{\textup{Sparse DETR.}}}
The number of object queries is 300 and we only use 10\% 
of the
encoder tokens on all benchmarks. We followed the original paper~\cite{roh2021sparse} for the other hyperparameters. Each adaptation step was conducted with the learning rate equal to the final learning rate of the prior step.

\paragraph{\textbf{\textup{OICR and CASD.}}}
We followed the original papers~\cite{tang2017multiple,huang2020comprehensive} for the hyperparameters.

\paragraph{\textbf{\textup{Object Proposals.}}}
The number of instances for each class in SyntheticPizza10 $\rightarrow$ RealPizza10 and Clipart1K $\rightarrow$ VOC2007 is unbalanced and the statistics information is reported in \autoref{tab:num_instance_voc} and \autoref{tab:num_instance_pizza}, respectively. The statistics information of object proposals mentioned in the main paper is shown in \autoref{tab:proposal}. 

\begin{table*}[hbt!]
\centering
\caption{Statistics of Clipart1k $\rightarrow$ VOC2007 (train+test): number of images ($\#\text{img}$), number of instances ($\#\text{ins}$), and relative size of human-labeled object instances on average ($\%\text{size}$).
}  
\label{tab:num_instance_voc}
\footnotesize
\resizebox{2.\columnwidth}{!}{
\begin{tabular}{l c c c c c c }
\toprule
\multirow{2}{*}{Name} & \multicolumn{3}{c}{Clipart} & \multicolumn{3}{c}{VOC}  \\
\cmidrule{2-4} \cmidrule{5-7} 
& $\#\text{img}$ & $\#\text{ins}$ & $\%\text{size}$ & $\#\text{img}$ & $\#\text{ins}$ & $\%\text{size}$ \\
\midrule
Aero & 41 & 73 & 14.7 & 442 & 591 & 26.3  \\
Bike & 27 & 36 & 20.1 & 482 & 690 & 22.0 \\
Bird & 135 & 265 & 9.4 & 612 & 945 & 20.2 \\
Boat & 88 & 129 & 12.1 & 353 & 553 & 17.2 \\
Bottle & 60 & 121 & 3.5 & 456 & 974 & 5.8 \\
Bus & 20 & 21 & 31.7 & 360 & 442 & 28.9 \\
Car & 103 & 202 & 11.0 & 1434 & 2451 & 19.8 \\
Cat & 43 & 50 & 8.4 & 659 & 734 & 41.9 \\
Chair & 181 & 340 & 8.6 & 862 & 1554 & 12.2 \\
Cow & 30 & 46 & 19.1 & 268 & 503 & 18.4 \\
\midrule
Total &&&&&&  \\
\bottomrule
\end{tabular}
\begin{tabular}{l c c c c c c}
\toprule
\multirow{2}{*}{Name} & \multicolumn{3}{c}{Clipart} & \multicolumn{3}{c}{VOC} \\
\cmidrule{2-4} \cmidrule{5-7} 
& $\#\text{img}$ & $\#\text{ins}$ & $\%\text{size}$ & $\#\text{img}$ & $\#\text{ins}$ & $\%\text{size}$ \\
\midrule
Table & 106 & 115 & 19.7 & 390 & 421 & 33.3 \\
Dog & 51 & 54 & 9.2 & 839 & 999 & 34.1\\
Horse & 46 & 79 & 17.4 & 561 & 710 & 30.2\\
Mbike & 16 & 17 & 51.2 & 467 & 664 & 28.3\\
Person & 521 & 1185 & 14.5 & 4015 & 9218 & 16.5\\
Plant & 100 & 178 & 5.6 & 469 & 994 & 11.6\\
Sheep & 27 & 76 & 8.1 & 193 & 499 & 13.0\\
Sofa & 42 & 52 & 19.1 & 452 & 487 & 35.6\\
Train & 45& 46 & 40.3 & 520 & 579 & 36.9\\
Tv & 65 & 80 & 8.4 & 485 & 632 & 12.8\\
\midrule
 & 1000 & 3165 & 12.9 & 9963 & 24640 & 19.9 \\
\bottomrule
\end{tabular}
}
\end{table*}

\begin{table*}[t]
\centering
\caption{Statistics of SyntheticPizza10 (train+test) $\rightarrow$ RealPizza10 (only test set is annotated, so here we report test statistic): number of images ($\#\text{img}$), number of instances ($\#\text{ins}$), and relative size of human-labeled object instances on average ($\%\text{size}$).
}
\label{tab:num_instance_pizza}
\footnotesize
\resizebox{2.\columnwidth}{!}{
\begin{tabular}{l c c c c c c c c c c cc}
\toprule
\multicolumn{2}{c}{Dataset} &  Pepperoni & Mushroom & Pepper & Olive & Basil & Bacon & Broccoli & Pineapple & Tomato & Onion & Total\\
\midrule
\multirow{3}{*}{Synthetic} & $\#\text{img}$ & 3741 & 3894 & 3901 & 3729 & 3868 & 3725 & 3904 & 3638 & 3979 & 3931 & 16340 
 \\
\cline{2-13}
& $\#\text{ins}$ &  29113 & 31723 & 29336 & 29001 & 36846 & 36113 & 31853 & 28675 & 31374 & 29561 & 313595
 \\
\cline{2-13}
 & $\%\text{size}$ & 2.4 & 2.2 & 2.4 & 2.4 & 1.4 & 1.3 & 2.2 & 2.3 & 2.3 & 2.7 & 2.1 
 \\
\midrule
\multirow{3}{*}{Real} & $\#\text{img}$ & 197 & 96 & 94 & 70 & 152 & 24 & 10 & 8 & 161 & 76 & 552
\\
\cline{2-13}
& $\#\text{ins}$ & 3638 & 1455 & 1239 & 841 & 1145 & 289 & 110 & 152 & 1697 & 706 & 11272
\\
\cline{2-13}
& $\%\text{size}$ & 1.3 & 1.0 & 1.0 & 0.6 & 1.4 & 1.8 & 1.5 & 0.9 & 1.3 & 1.2 & 1.2
\\
\bottomrule
\end{tabular}
}
\end{table*}

\begin{table*}[hbt!]
\centering
\caption{Statistical information of the number of object proposals per image generated by the Faster R-CNN backbone in source domains.}
\label{tab:proposal}
\footnotesize
\begin{tabular}{l c c c c }
\toprule
Datasets & Min & Max & Mean & Std \\
\midrule
PASCAL VOC 2007 train & 300.0 & 1800.0 & 437.3 & 195.5 \\
PASCAL VOC 2007 test & 300.0 & 1500.0 & 424.8 & 185.0 \\
RealPizza10 2007 train & 300.0 & 1800.0 & 441.0 & 225.7 \\
RealPizza10 2007 test & 300.0 & 1500.0 & 482.6 & 236.7 \\
\bottomrule
\end{tabular}
\end{table*}

\paragraph{\textbf{\textup{Cost of Training and Computing Resources.}}}
We train \DFOSDA based on Faster R-CNN on a Tesla K80 GPU. As shown in~\autoref{tab:training time}, we have the following observations. (1) The warm-up stage takes much less time than the main stage, indicating our domain adaptation to be lightweight. (2) Standard CycleGAN training brings in the most additional computation overhead. 
\begin{table*}[hbt!]
\centering
\caption{Training time for different stages on Clipart1K $\rightarrow$ VOC2007 (Clip $\rightarrow$ VOC), SyntheticPizza10 $\rightarrow$ RealPizza10 (SPizza $\rightarrow$ RPizza),  Watercolor2K $\rightarrow$ VOC2007-sub (Water $\rightarrow$ VocS), and Comic2K $\rightarrow$ VOC2007-sub (Comi $\rightarrow$ VocS) based on Faster R-CNN FSOD
backbone. `-' denotes that we run experiments without the 3rd adaptation step FSOD-3, since VOC2007 and MS-COCO lack clean background.}
\label{tab:training time}
\footnotesize
\resizebox{1.9\columnwidth}{!}{
\footnotesize
\begin{tabular}{c|c|c|c|c|c|c|c|c }
\multicolumn{1}{c|}{Training time} & \multicolumn{1}{c|}{Dataset} & \multicolumn{5}{c|}{Warm-Up Stage} & \multicolumn{1}{c|}{Main Stage} & \multicolumn{1}{c}{CycleGAN}\\
\hline
Stage & & FSOD-1 & FSOD-2 & FSOD-3 & FSOD-4 & FSOD-5 & CASD & \\
\hline
\multirow{4}{*}{hour} & SPizza $\rightarrow$ RPizza & 14 & 10 & 13 & 2 & 2 & 79 & 216 \\
\cline{2-9}
& Clip $\rightarrow$ Voc & 7 & 6 & - & 3 & 3 & 102 & 144 \\
\cline{2-9}
& Water $\rightarrow$ VocS & 8 & 8 & - & 2 & 2 & 81 & 144\\
\cline{2-9}
& Comi $\rightarrow$ VocS & 4 & 4 & - & 8 & 8 & 81 & 144 \\
\end{tabular}
}
\vspace{-3mm}
\end{table*}

\section{Additional Main Results}
In \autoref{sec:main_results} of the main paper we show the main results based on mAP values. Here, we list the whole mAP with per class AP values. \autoref{tab:voc_baseline} and \autoref{tab:pizza_baseline} summarize the detection results on Clipart1K $\rightarrow$ VOC2007 and SyntheticPizza10 $\rightarrow$ RealPizza10 based on Faster R-CNN FSOD backbone, respectively. {\DFOSDA} incorporated with OICR is denoted as ${\DFOSDA}_\text{oicr}$, with CASD is denoted as ${\DFOSDA}_\text{casd}$, and with CASD+W2N is denoted as ${\DFOSDA}_\text{casd+w2n}$. The results of our warm-up stage are denoted as ${\DFOSDA}_\text{warm-up}$. 

The detection performance does not benefit from using OICR or CASD twice (once for proposal and once for object detection), since doing so does not improve the generated proposals. The result of training CASD two times is denoted as CASD$^2$. As shown in~\autoref{tab:voc_baseline}, on Clipart1K $\rightarrow$ VOC2007, ${\DFOSDA}_\text{casd}$ reaches $64.8\%$ mAP, outperforming the original CASD by $7.8\%$ mAP, while CASD$^2$ reaches $57.4\%$ mAP, outperforming the original CASD by $0.4\%$ mAP. The detection result of using OICR twice will be the same as using OICR once, because OICR directly uses the proposals produced by selective search without refinement on their bounding boxes. ${\DFOSDA}_\text{casd+w2n}$ reaches $66.9\%$ mAP, outperforming the original CASD+W2N by $1.5\%$ mAP. This further validates that {\em \DFOSDA is a general framework that can be combined with different WSOD methods} to improve their object proposal generation and thus overall performance.

\begin{table*}[hbt!]
\centering
\caption{Results (AP in $\%$) for different methods on Clipart1K $\rightarrow$ VOC2007 and SyntheticPizza10 $\rightarrow$ RealPizza10. We denote as Upper-Bound the FSOD (Faster R-CNN or Sparse DETR) results, trained and tested on {\em fully-annotated} target domain to indicate the weak upper-bound performance of our methods. Our warm-up stage is compared with CD models and our main stage is compared with SD models. The upper part shows the results using CD models. The lower part shows the results using SD methods.
Faster R-CNN in CD means we trained our network on fully-annotated source and test on fully-annotated target domains. The best and second best results for {\DFOSDA} compared with baselines are shown in red and blue.}
\vspace{-3mm}
\label{tab:voc_pizza_baseline}
\begin{subtable}[h]{\textwidth}
\caption{Clipart1K $\rightarrow$ VOC2007.}
\vspace{-2mm}
\label{tab:voc_baseline}
\footnotesize
\resizebox{\columnwidth}{!}{
\begin{tabular}{c|c|c |c c c c c c c c c c c c c c c c c c c c c c c c c c c c c c c c c c c c c c c c|}
Type & Method & mAP & Aero & Bike & Bird & Boat & Bottle & Bus & Car & Cat & Chair & Cow & Table & Dog & Horse & Mbike & Person & Plant & Sheep & Sofa & Train & Tv\\
\hline
\hline
\multirow{1}{*}{}& Upper-Bound ~\cite{ren2015faster} & \em 69.9 & \em 69.8 & \em 79.1 & \em 67.5 & \em 56.5 & \em 54.4 & \em 77.2 &
\em 82.2 & \em 80.9 & \em 50.1 & \em 78.4 & \em 64.5 & \em 78.4 & \em 83.7 & \em 72.3 & \em 77.2 & \em 38.3 & \em 70.9 & \em 66.6 & \em 77.5 & \em 71.9 \\
\hline
\hline
\multirow{3}{*}{\texttt{CD}} & Faster R-CNN~\cite{ren2015faster} & 22.8 & 10.7 & 39.7 & 30.5 & 8.6 & 19.3 & 27.4 & 48.0 & 4.5 & 23.7 & 21.2 & 7.9 & 19.0 & 21.9 & 21.5 & 45.0 & 17.4 & 16.1 & 22.5 & 25.5 & 25.0 \\
\cline{2-23}
& DT+PL~\cite{inoue2018cross} & 34.6 & 18.8 & 55.8 & 33.2 & 20.4 & 18.8 & 47.2 & 56.2
& 15.8 & 27.4 & 45.5 & 10.7 & 25.9 & 54.1 & 54.3 & 47.6 & 10.6 & 35.4 & 42.3 & 47.0 &25.9 \\
\cline{2-23}
& PADOD~\cite{hsu2020progressive} & 24.2 & 13.3 & 40.3 & 28.8 & 12.6 & 20.2 & 32.2 & 46.7 & 7.5 & 25.9 & 24.0 & 13.8 & 19.3 & 21.2 & 17.5 & 44.0 & 17.6 & 17.2 & 24.4 & 28.5 & 29.4\\
\hline
\multirow{1}{*}{\texttt{Ours}} & ${\DFOSDA}_\text{warm-up}$ & 37.3 & 20.7 & 60.8 & 37.2 & 19.4 & 25.0 & 51.1 & 59.7 & 17.2 & 30.4 & 44.4 & 17.3 & 27.4 & 55.8 & 56.8 & 47.6 & 12.9 & 38.4 & 45.5 & 50.3 & 27.6\\
\hline
\hline
\multirow{12}{*}{\texttt{SD}} & WSDDN~\cite{bilen2016weakly} & 34.8 & 39.4 & 50.1 & 31.5 & 16.3 & 12.6 & 64.5 & 42.8 & 42.6 & 10.1 & 35.7 & 24.9 & 38.2 & 34.4 & 55.6 & 9.4 & 14.7 & 30.2 & 40.7 & 54.7 & 46.9 \\
\cline{2-23}
& OICR~\cite{tang2017multiple} & 41.2 & 58.0 & 62.4 & 31.1 & 19.4 & 13.0 & 65.1 & 62.2 & 28.4 & 24.8 & 44.7 & 30.6 & 25.3 & 37.8 & 65.5 & 15.7 & 24.1 & 41.7 & 46.9 & 64.3 & 62.6 \\
\cline{2-23}
& PCL~\cite{tang2018pcl} & 43.5 & 54.4 & 69.0 & 39.3 & 19.2 & 15.7 & 62.9 & 64.4 & 30.0 & 25.1 & 52.5 & 44.4 & 19.6 & 39.3 & 67.7 & 17.8 & 22.9 & 46.6 & 57.5 & 58.6 & 63.0 \\
\cline{2-23}
& WeakRPN~\cite{tang2018weakly} & 45.3 & 57.9 & 70.5 & 37.8 & 5.7 & 21.0 & 66.1 & 69.2 & 59.4 & 3.4 & 57.1 & \color[rgb]{1,0,0}57.3 & 35.2 & 64.2 & 68.6 & 32.8 & 28.6 & 50.8 & 49.5 & 41.1 & 30.0\\
\cline{2-23}
& C-MIL~\cite{wan2019c} & 50.5 & 62.5 & 58.4 & 49.5 & 32.1 & 19.8 & 70.5 & 66.1 & 63.4 & 20.0 & 60.5 & 52.9 & 53.5 & 57.4 & 68.9 & 8.4 & 24.6 & 51.8 & 58.7 & 66.7 & 63.5 \\
\cline{2-23}
& WSOD2(+Reg)~\cite{zeng2019wsod2} & 53.6 & 65.1 & 64.8 & 57.2 & 39.2 & 24.3 & 69.8 & 66.2 & 61.0 & 29.8 & 64.6 & 42.5 & 60.1 & 71.2 & 70.7 & 21.9 & 28.1 & 58.6 & 59.7 & 52.2 & 64.8 \\
\cline{2-23}
& Pred Net~\cite{arun2019dissimilarity} & 52.9 & 66.7 & 69.5 & 52.8 & 31.4 & 24.7 & \color[rgb]{0,0,1}74.5 & 74.1 & 67.3 & 14.6 & 53.0 & 46.1 & 52.9 & 69.9 & 70.8 & 18.5 & 28.4 & 54.6 & \color[rgb]{0,0,1}60.7 & 67.1 & 60.4  \\
\cline{2-23}
& C-MIDN~\cite{gao2019c} & 52.6 & 53.3 & 71.5 & 49.8 & 26.1 & 20.3 & 70.3 & 69.9 & 68.3 & 28.7 & 65.3 & 45.1 & 64.6 & 58.0 & 71.2 & 20.0 & 27.5 & 54.9 & 54.9 & 69.4 & 63.5 \\
\cline{2-23}
& MIST(+Reg)~\cite{ren2020instance} & 54.9 & \color[rgb]{0,0,1}68.8 & \color[rgb]{0,0,1}77.7 & 57.0 & 27.7 & 28.9 & 69.1 &74.5 & 67.0 & 32.1 & 73.2 & 48.1 & 45.2 & 54.4 &73.7 & 35.0 & 29.3 & 64.1 & 53.8 & 65.3 & 65.2 \\
\cline{2-23}
& CASD~\cite{huang2020comprehensive} & 57.0 & 67.2 & 71.5 & 57.8 & 41.5 & 23.4 & 72.9 & 70.3 & \color[rgb]{0,0,1}75.5 & 21.5 & 64.8 & \color[rgb]{0,0,1}53.8 & 71.8 & 65.0 & 72.5 & 32.6 & 25.0 & 56.6 & 58.5 & \color[rgb]{0,0,1}69.5 & \color[rgb]{0,0,1}68.2\\
\cline{2-23}
& CASD$^2$ & 57.4 & 56.1 & 64.6 & 61.4 & \color[rgb]{1,0,0}60.9 & 35.1 & 59.9 & 76.9 & 56.6 & 27.8 & 73.6 & 51.2 & 60.1 & 70.5 & 72.5 & 34.4 & 54.0 & 70.5 & 45.1 & 53.1 & 63.7\\
\cline{2-23}
& CASD+W2N~\cite{huang2022w2n} & \color[rgb]{0,0,1}65.4 & \color[rgb]{1,0,0}74.0 & \color[rgb]{1,0,0} 81.7 & \color[rgb]{0,0,1}71.2 & 48.9 & 51.0 & \color[rgb]{1,0,0}78.6 & 82.3 & \color[rgb]{1,0,0}83.5 & 29.1 & 76.9 & 51.5 & \color[rgb]{1,0,0}82.1 & 76.9 & \color[rgb]{1,0,0}79.1 & 28.5 & 34.3 & 65.0 & \color[rgb]{1,0,0}64.2 & \color[rgb]{1,0,0}75.2 & \color[rgb]{1,0,0}74.8\\
\hline
\multirow{2}{*}{\texttt{Ours}}& 
${\DFOSDA}_\text{casd}$ & 64.8 & 62.7 & 64.9 &69.9 &47.9 & \color[rgb]{0,0,1}57.9 & 74.3 &	\color[rgb]{0,0,1}85.7 & 59.6 & \color[rgb]{1,0,0}43.4 & \color[rgb]{0,0,1}82.2 & 39.6 & 67.2 & \color[rgb]{0,0,1}84.0 & \color[rgb]{0,0,1}77.8 & \color[rgb]{0,0,1}74.0 & \color[rgb]{0,0,1}50.6 & \color[rgb]{1,0,0}74.6 & 48.8 & 66.7 & 64.6 \\
\cline{2-23}
& ${\DFOSDA}_\text{casd+w2n}$ & \color[rgb]{1,0,0}66.9 & 58.6 & 69.1	& \color[rgb]{1,0,0}77.9	& \color[rgb]{0,0,1}49.3	& \color[rgb]{1,0,0}78.1	& 73.2	& \color[rgb]{1,0,0}89.0	& 64.9 & \color[rgb]{0,0,1}39.6	& \color[rgb]{1,0,0}83.5 & 33.0	& \color[rgb]{0,0,1}77.7 & \color[rgb]{1,0,0}95.2 & 77.0 & \color[rgb]{1,0,0}75.9 & \color[rgb]{1,0,0}50.7 & \color[rgb]{0,0,1}74.4	& 44.8 & 66.6 & 61.0
\end{tabular}
}
\end{subtable}
\newline
\begin{subtable}[h]{\textwidth}
\caption{SyntheticPizza10 $\rightarrow$ RealPizza10.}
\vspace{-2mm}
\label{tab:pizza_baseline}
\resizebox{\columnwidth}{!}{
\footnotesize
\begin{tabular}{c|c|c |c c c c c c c c c c c|}
Type & Method & mAP & Pepperoni & Mushroom & Pepper & Olive & Basil & Bacon & Broccoli & Pineapple & Tomato & Onion\\
\hline
\hline
\multirow{3}{*}{\texttt{CD}} & Faster R-CNN~\cite{ren2015faster} & 4.3 & 12.1 & 0.4 & 9.6 & 5.0 & 3.4 & 0.3 & 1.0 & 1.0 & 9.7 & 0.9 \\
\cline{2-13}
& DT+PL~\cite{inoue2018cross}& 14.9 & 30.7 & 4.3 & 11.6 & 25.3 & \color[rgb]{0,0,1}42.7 & 1.3 & 3.6 & 2.4 & 21.4 & 5.2\\
\cline{2-13}
& PADOD~\cite{hsu2020progressive} & 8.1 & 19.5 & 0.2 & 3.4 & 11.8 & 30.2 & 0.2 & 1.1 & 0.5 & 13.3 & 0.8
\\
\hline
\texttt{Ours}
& ${\DFOSDA}_\text{warm-up}$ & 
\color[rgb]{0,0,1}17.9 & \color[rgb]{0,0,1}31.0 & 8.3 & 11.8 & \color[rgb]{1,0,0}28.1 & \color[rgb]{1,0,0}45.5 & 0.8 & 9.8 & 9.8 & \color[rgb]{0,0,1}21.5 & \color[rgb]{0,0,1}12.7
 \\
\hline
\hline
\multirow{2}{*}{\texttt{SD}} & OICR~\cite{tang2017multiple} &  4.7
 & 0.2 & 1.3 & 4.5 & 0.1 & 0 & 8.8 & \color[rgb]{1,0,0}19.4 & \color[rgb]{0,0,1}11.0 & 1.0 & 0.8\\
\cline{2-13}
& CASD~\cite{huang2020comprehensive} &  12.9 & 12.7 & \color[rgb]{0,0,1}19.5 & \color[rgb]{0,0,1}14.8 & 10.5 & 13.7 & \color[rgb]{0,0,1}10.4 & 10.1 & \color[rgb]{1,0,0}14.5 & 11.7 & 10.7 \\
\hline
\texttt{Ours} 
& ${\DFOSDA}_\text{casd}$ & 
\color[rgb]{1,0,0}25.1 & \color[rgb]{1,0,0}43.9 & \color[rgb]{1,0,0}35.1 & \color[rgb]{1,0,0}14.9 & \color[rgb]{0,0,1}27.3 & 41.8 & \color[rgb]{1,0,0}9.2 & \color[rgb]{0,0,1}12.5 & 8.5 & \color[rgb]{1,0,0}28.4 & \color[rgb]{1,0,0}29.2 \\
\end{tabular}
}
\end{subtable}
\vspace{-3mm}
\end{table*}
\section{Additional Ablation Study}
\label{sec:add_ablation}
In \autoref{sec:ablation_study} of the main paper we show the ablation study results based on the mAP values. Here, we list the whole mAP values with per class AP values.
\begin{itemize}
    \item \textbf{Effectiveness of Progressive Adaptation.} As shown in~\autoref{tab:pizza_ef}, each adaptation step in our warm-up stage is not only helpful in terms of mAP, but it also benefits for each class.
    \item \textbf{Impact of Adaptation Order.}~\autoref{tab:pizza_order} shows that our {\em progressive} adaptation order that gradually reduces domain gap achieves the best performance -- it is better to first fine-tune the FSOD on intermediate images or pseudo-labeled images, and then fine-tune on the augmented images.
    \item \textbf{Generalizability of the Warm-up Stage across FSODs.} As shown in~\autoref{tab:pizza_ge}, our ${\DFOSDA}_\text{warm-up}$ and ${\DFOSDA}_\text{casd}$ based on Sparse DETR yield 0.6$\%$ and 1.1$\%$ improvement in terms of mAP, respectively, compared with Faster R-CNN backbone. 
    \item \textbf{Main Stage Configurations.} As shown in~\autoref{tab:main_all}, two key components of our method are both effective and complementary to each other. Note that the current transformer-based detectors rely on relatively large amounts of annotate data; therefore, we found that it was difficult to train Sparse DETR on Clipart1K with only 500 training images; by contrast, Sparse DETR worked reasonably well on SyntheticPizza10 with 14,802 training images. Accordingly, for the experiment on the Clipart1K $\rightarrow$ VOC2007 datasets, we mainly focus on Faster R-CNN. We leave the investigation of Sparse DETR on Clipart1K $\rightarrow$ VOC2007 as interesting future work, by either exploring additional synthetic data to increase the synthetic training dataset size or leveraging more data-efficient transformer-based detectors.
    \item \textbf{Identifying Object Detection Errors.} We use TIDE~\cite{bolya2020tide} to analyse the {\color{blue}{classification}}, {\color{brown}{localization}}, {\color{green}{both Cls and Loc}}, {\color{pink}{duplicate detection}}, {\color{violet}{background}}, and {\color{red}{missed GT}} errors in DT+PL, CASD, and our model. Each chart shows the relative percentage of each type of error. As shown in~\autoref{fig:error2}, \DFOSDA effectively reduces the localization error compared with other two baselines. Here classification error indicates object localized correctly but misclassified; localization error indicates object classified correctly but mislocalized; both Cls and Loc error indicates object misclassified and mislocalized; duplicate detection error indicates object matched with a GT which has already matched with another higher confidence scoring prediction; background error indicates background detected as foreground; missed GT error indicates ground-truth that not matched with any predictions.  

    \begin{figure}
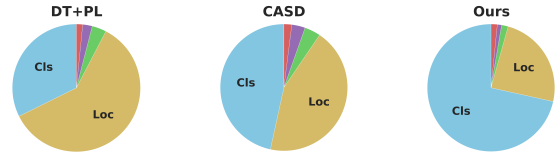

        \centering
        \begin{minipage}{\linewidth}
      \begin{minipage}{0.32 \linewidth}
        \centering
        \includegraphics[trim={0 53cm 5cm 0cm},clip,width = .7\linewidth]{images/DT+PL_bbox_summary.png}
        \end{minipage}
      \begin{minipage}{0.32 \linewidth}
        \centering
        \includegraphics[trim={0 53cm 5cm 0cm},clip,width = .7\linewidth]{images/CASD_bbox_summary_v3.png}
      \end{minipage}
      \begin{minipage}{0.32 \linewidth}
        \centering
        \includegraphics[trim={0 53cm 5cm 0cm},clip,width = .7\linewidth]{images/Ours_bbox_summary_v2.png}
      \end{minipage}
      \hfill
    \end{minipage}
        \caption{Summary of errors on DT+PL, CASD and our method.}
        \label{fig:error2}
    \end{figure}
\end{itemize}

\begin{table*}[hbt!]
    \centering
    \caption{Effectiveness of progressive adaptation: each adaptation step in our warm-up stage is helpful not only in terms of mAP, but also benefits for each class.}
    \label{tab:pizza_ef}
    \footnotesize
    \resizebox{2.0\columnwidth}{!}{
    \begin{tabular}{c@{\hspace{5mm}} c@{\hspace{5mm}}  c@{\hspace{2mm}} c@{\hspace{2mm}} c@{\hspace{2mm}} c@{\hspace{2mm}} c@{\hspace{2mm}} c@{\hspace{2mm}} c@{\hspace{2mm}} c@{\hspace{2mm}} c@{\hspace{2mm}} c@{\hspace{2mm}} c@{\hspace{2mm}} c@{\hspace{2mm}} }
    \toprule
    FSOD & Step & mAP & Pepperoni & Mushroom & Pepper & Olive & Basil & Bacon & Broccoli & Pineapple & Tomato & Onion\\
    \midrule
    \multirow{5}{*}{\texttt{Faster R-CNN}} & FSOD-1 & 4.3 & 12.1 & 0.4 & 9.6 & 5.0 & 3.4 & 0.3 & 1.0 & 1.0 & 9.7 & 0.9 \\
    \cline{2-13}
    & FSOD-2 & 9.7 & 22.3 & 0.8 & 9.6 & 14.5 & 33.2 & 0.8 & 0.7 & 0.8 & 13.4 & 1.0 \\
    \cline{2-13}
    & FSOD-3 & 10.3 & 23.0 & 1.0 & 10.1 & 15.4 & 35.7 & 0.8 & 0.8 & 0.6 & 14.4 & 1.1 \\
    \cline{2-13}
    & FSOD-4 & 15.0 & 28.7 & 4.6 & 11.0 & 22.2 & 39.9 & 9.3 & 1.8 & 9.6 & 16.6 & 6.3 \\
    \cline{2-13}
    & FSOD-5 & 17.9 & 31.0 & 8.3 & 11.8 & 28.1 & 45.5 & 0.8 & 9.8 & 9.8 & 21.5 & 12.7 \\
    \midrule
    \multirow{5}{*}{\texttt{Sparse DETR}} & FSOD-1 & 3.7 & 10.0 & 0.2 &	0.8 & 20.2 & 0.7 & 0.2 & 0.4 & 0.6 & 1.0 & 2.6 \\
    \cline{2-13}
    & FSOD-2 & 9 & 27.0 & 1.7 & 0.6 & 	24.5 & 19.1 &	0.1 &	2.3 & 2.5 & 8.3 & 4.0 \\
    \cline{2-13}
    & FSOD-3 & 10.3	 & 30.40 & 0.9 & 1.2 & 28.1 & 27.7 &	0.0 & 1.2 &	1.3 & 9.2 & 2.5\\
    \cline{2-13}
    & FSOD-4 & 18.2	& 47.7 & 4.7 & 5.9 & 39.8 &	46.0 & 0.0 & 1.7 &	3.1 &	25.1 &	7.8 \\
    \cline{2-13}
    & FSOD-5 & 18.5 & 48.3 & 6.0 & 5.3 & 43.0 & 43.1 & 0.1 & 1.4 & 2.4 & 26.1 & 9.2 \\
    \bottomrule
    \end{tabular}
    }
\end{table*}

\begin{table*}[hbt!]
    \centering
    \caption{Impact of adaptation order (Faster R-CNN backbone).}
    \label{tab:pizza_order}
    \footnotesize
    \resizebox{2.0\columnwidth}{!}{
    \begin{tabular}{c@{\hspace{5mm}}c@{\hspace{5mm}}c@{\hspace{2mm}} c@{\hspace{2mm}} c@{\hspace{2mm}} c@{\hspace{2mm}} c@{\hspace{2mm}} c@{\hspace{2mm}} c@{\hspace{2mm}} c@{\hspace{2mm}} c@{\hspace{2mm}} c@{\hspace{2mm}} c@{\hspace{2mm}}}
    \toprule
    Adaptation Domain & mAP & Pepperoni & Mushroom & Pepper & Olive & Basil & Bacon & Broccoli & Pineapple & Tomato & Onion\\
    \midrule
    $\mathcal{S}$  & 4.3 & 12.1 & 0.4 & 9.6 & 5.0 & 3.4 & 0.3 & 1.0 & 1.0 & 9.7 & 0.9 \\
    \midrule
    $\mathcal{S}$ $\rightarrow$ $\mathcal{G}_1$  $\rightarrow$ $\mathcal{G}_2$ & 10.3 & 23.0 & 1.0 & 10.1 & 15.4 & 35.7 & 0.8 & 0.8 & 0.6 & 14.4 & 1.1 \\
    \midrule
    $\mathcal{S}$ $\rightarrow$ $\mathcal{G}_2$  $\rightarrow$ $\mathcal{G}_1$ & 5.9 & 5.8 & 1.1 & 9.7 & 6.9 & 20.4 & 0 & 2.3 & 0.5 & 9.1 & 3.2\\
    \midrule
    $\mathcal{G}_2$ $\rightarrow$ $\mathcal{T}$  $\rightarrow$ Aug.$\mathcal{T}$  & 17.9 & 31.0 & 8.3 & 11.8 & 28.1 & 45.5 & 0.8 & 9.8 & 9.8 & 21.5 & 12.7 \\
    \midrule
    $\mathcal{G}_2$ $\rightarrow$ Aug.$\mathcal{T}$  $\rightarrow$ $\mathcal{T}$  & 17.3 & 32.6 & 6.3 & 8.3 & 28.4 & 47.3 & 0.6 & 4.9 & 10.4 & 22.4 & 11.3\\
    \bottomrule
    \end{tabular}
    }
\end{table*}

\begin{table*}[hbt!]
    \centering
    \caption{Generalizability of the warm-up stage across FSODs.}
    \label{tab:pizza_ge}
    \footnotesize
    \resizebox{2.0\columnwidth}{!}{
    \begin{tabular}{c@{\hspace{5mm}}c@{\hspace{5mm}} c@{\hspace{2mm}}c@{\hspace{2mm}} c@{\hspace{2mm}} c@{\hspace{2mm}} c@{\hspace{2mm}} c@{\hspace{2mm}} c@{\hspace{2mm}} c@{\hspace{2mm}} c@{\hspace{2mm}} c@{\hspace{2mm}} c@{\hspace{2mm}} c@{\hspace{2mm}}}
    \toprule
    FSOD & Stage & mAP & Pepperoni & Mushroom & Pepper & Olive & Basil & Bacon & Broccoli & Pineapple & Tomato & Onion\\
    \midrule
    \multirow{2}{*}{\texttt{Faster R-CNN}} & ${\DFOSDA}_\text{warm-up}$ & 17.9 & 31.0 & 8.3 & 11.8 & 28.1 & 45.5 & 0.8 & 9.8 & 9.8 & 21.5 & 12.7 \\
    \cline{2-13}
    & ${\DFOSDA}_\text{casd}$ & 
    25.1 & 43.9 & 35.1 & 14.9 & 27.3 & 41.8 & 9.2 & 12.5 & 8.5 & 28.4 & 29.2 \\
    \midrule
    \multirow{2}{*}{\texttt{Sparse DETR}} & ${\DFOSDA}_\text{warm-up}$ & 18.5 & 48.3 & 6.0 & 5.3 & 43.0 & 43.1 & 0.1 & 1.4 & 2.4 & 26.1 & 9.2 \\
    \cline{2-13}
    & ${\DFOSDA}_\text{casd}$ & 
    26.2 & 51.9 & 35.5 & 18.9 & 33.1 & 47.4 & 11.2 & 9.6 & 5.4 & 29.3 & 20.1\\
    \bottomrule
    \end{tabular}
    }
\end{table*}

\begin{table*}[hbt!]
\centering
\caption{Ablation study of {\DFOSDA} main configurations on (a) Clipart1K $\rightarrow$ VOC2007 (Faster R-CNN backbone), (b)  SyntheticPizza10 $\rightarrow$ RealPizza10  (Faster R-CNN backbone) and (c) SyntheticPizza10 $\rightarrow$ RealPizza10  (Sparse DETR backbone). ``\textbf{FE}'' and ``\textbf{OP}'' denote the domain specific pre-trained feature extractor and weakly-supervised object proposal generator, respectively.}\label{tab:main_all}

\begin{subtable}[h]{\textwidth}
\caption{Clipart1K $\rightarrow$ VOC2007 (Faster R-CNN backbone).}
\label{tab:voc_ab_f}
\footnotesize
\resizebox{\columnwidth}{!}{
\begin{tabular}{c@{\hspace{2mm}}c@{\hspace{2mm}}c @{\hspace{2mm}}c c c c c c c c c c c c c c c c c c c c c c c c c c c c c c c c c c c c c c c c}
\toprule
Type & Method & mAP & aero & bike & bird & boat & bottle & bus & car & cat & chair & cow & table & dog & horse & mbike & person & plant & sheep & sofa & train & tv\\
\midrule
SD & OICR & 41.2 & 58.0 & 62.4 & 31.1 & 19.4 & 13.0 & 65.1 & 62.2 & 28.4 & 24.8 & 44.7 & 30.6 & 25.3 & 37.8 & 65.5 & 15.7 & 24.1 & 41.7 & 46.9 & 64.3 & 62.6
\\
\midrule
\multirow{3}{*}{${\DFOSDA}_\text{oicr}$} & \textbf{+FE} & 44.7 & 53.8 & 52.5 &	41.1 & 37.4 & 27.8 & 53.9 & 63.5 & 39.1 & 30.5 & 59.5 & 40.7 & 42.6 & 47.6 & 52.1 & 23.5 & 36.1 & 55.9 & 40.0 & 45.1 & 50.7
\\
\cline{2-23}
& \textbf{+OP} & 47.2 & 23.4 & 54.4 & 46.9 & 34.6 &	46.5 & 69.4 & 78.0 & 10.1 & 44.7 & 65.6 & 27.7 & 25.9 & 52.8 & 64.3 & 65.3 & 32.5 & 54.7 & 42.1 & 52.8 & 52.3\\
\cline{2-23}
& \textbf{+FE+OP} & 52.7 & 39.1 & 60.6 & 56.2 & 37.4 & 48.0 & 67.8 & 81.0 & 18.6 &  51.8 & 67.5 & 38.1 & 31.3 & 72.0 & 67.8 & 70.2 & 40.0 & 60.6 & 40.9 & 56.4 & 49.3\\
\midrule
SD & CASD & 57.0 & 67.2 & 71.5 & 57.8 & 41.5 & 23.4 & 72.9 & 70.3 & 75.5 & 21.5 & 64.8 & 53.8 & 71.8 & 65.0 & 72.5 & 32.6 & 25.0 & 56.6 & 58.5 & 69.5 & 68.2 \\
\midrule
\multirow{3}{*}{${\DFOSDA}_\text{casd}$} & \textbf{+FE} & 60.0 & 
51.0 & 71.1 & 72.1 & 38.1 & 27.5 & 76.1 & 71.6 & 74.0 & 27.2 & 64.3 & 58.8 & 81.6 & 88.1 & 71.0 & 63.1 & 19.5 & 53.2 & 58.7 & 69.4 & 64.0\\
\cline{2-23}
& \textbf{+OP} & 60.1 & 38.3 & 67.6 & 63.2 & 45.4 & 62.0 & 77.7 & 88.9 & 24.1 & 56.3 & 76.9 & 44.9 & 41.4 &	76.8 & 77.4 & 75.1 & 42.2 & 68.6 & 52.4 & 62.2 & 61.4 \\  
\cline{2-23}
& \textbf{+FE+OP} & 64.8 & 62.7 & 64.9 & 69.9 & 47.9 &57.9 & 74.3 &	85.7 & 59.6 & 43.4 & 82.2 &	39.6 & 67.2 &84.0 &77.8 & 74.0 & 50.6 &74.6 & 48.8 & 66.7 & 64.6
\\
\bottomrule
\end{tabular}
}
\end{subtable}
\newline
\newline
\begin{subtable}[h]{\textwidth}
\caption{SyntheticPizza10 $\rightarrow$ RealPizza10  (Faster R-CNN backbone).}
\label{tab:pizza_ab_s}
\footnotesize
\resizebox{\columnwidth}{!}{
\begin{tabular}{c@{\hspace{5mm}}c@{\hspace{5mm}}c @{\hspace{5mm}}c @{\hspace{2mm}}c@{\hspace{2mm}}c@{\hspace{2mm}} c@{\hspace{2mm}} c@{\hspace{2mm}} c@{\hspace{2mm}} c@{\hspace{2mm}} c@{\hspace{2mm}} c@{\hspace{2mm}} c@{\hspace{2mm}} c}
\toprule
Type & Method & mAP & Pepperoni & Mushroom & Pepper & Olive & Basil & Bacon & Broccoli & Pineapple & Tomatoes & Onion\\
\midrule
\texttt{SD} & OICR &  4.7
 & 0.2 & 1.3 & 4.5 & 0.1 & 0 & 8.8 & 19.4 & 11.0 & 1.0 & 0.8 \\
\midrule
\multirow{3}{*}{${\DFOSDA}_\text{oicr}$} &
\textbf{+FE} &  8.5 & 4.4	&  12.5 & 12.2 &	7.2 & 6.1 &	7.4 & 8.6 &	13.2 & 5.1 & 7.9\\
\cline{2-13}
&\textbf{+OP} &  12.6 & 23.0 & 18.5 & 8.5 & 14.7 & 20.8 & 5.0 & 2.3 & 3.9 &	13.9 & 15.8 \\
\cline{2-13}
&\textbf{+FE+OP} &  13.8 & 24.3 & 19.7 & 10.0 & 15.2 & 21.9 & 3.7 & 7.5 & 3.6 &	16.3 & 15.4 \\
\midrule
\texttt{SD} & CASD & 12.9 & 12.7 & 19.5 & 14.8 & 10.5 & 13.7 & 10.4 & 10.1 & 14.5 & 11.7 & 10.7  \\
\midrule
\multirow{3}{*}{${\DFOSDA}_\text{casd}$} & \textbf{+FE} & 14.8 & 10.2 & 11.3 & 14.6 & 10.1 & 10.0 & 19.0 & 30.9 & 21.0 & 10.9 & 10.5 
 \\
\cline{2-13}
& \textbf{+OP} & 24.0 & 45.8 & 36.7 & 14.8 & 25.8 & 37.4 & 3.2 & 12.3 & 3.5 & 32.1 & 28.3\\
\cline{2-13}
& \textbf{+FE+OP} & 25.1 & 43.9 & 35.1 & 15.0 &	27.3 & 41.8 & 9.2 & 12.5 & 8.5 & 28.4 & 29.2 \\
\bottomrule
\end{tabular}
}

\end{subtable}
\newline
\newline
\begin{subtable}[h]{\textwidth}
\caption{SyntheticPizza10 $\rightarrow$ RealPizza10  (Sparse DETR backbone).}
\label{tab:pizza_ab_f}
\footnotesize
\resizebox{\columnwidth}{!}{
\begin{tabular}{c@{\hspace{5mm}}c@{\hspace{5mm}}c@{\hspace{5mm}}@{\hspace{2mm}}c@{\hspace{2mm}} c@{\hspace{2mm}} c@{\hspace{2mm}} c@{\hspace{2mm}} c@{\hspace{2mm}} c@{\hspace{2mm}} c@{\hspace{2mm}} c@{\hspace{2mm}} c@{\hspace{2mm}} c@{\hspace{2mm}} c}
\toprule
Type & Method & mAP & Pepperoni & Mushroom & Pepper & Olive & Basil & Bacon & Broccoli & Pineapple & Tomatoes & Onion\\
\midrule
\texttt{SD} & OICR &  5.9 & 12.1 & 9.9 & 2.8 & 5.4 & 12.7 & 1.0 & 0.3 & 0.1 & 5.5 & 9.4 \\
\midrule
\multirow{3}{*}{${\DFOSDA}_\text{oicr}$} &
\textbf{+FE} &  10.8 & 21.7 & 17.1 & 9.3 & 11.6 & 15.2 & 0.4 & 4.5 & 0.8 & 15.6 & 12.1\\
\cline{2-13}
&\textbf{+OP} & 13.4 & 22.1 & 16.8 & 9.0 & 13.5 & 23.7 & 10.9 & 5.5 & 2.2 & 15.6 & 14.3 \\
\cline{2-13}
&\textbf{+FE+OP} &  15.4 & 28.0 & 13.2 & 9.5 & 21.5 & 24.1 & 8.0 & 14.0 & 7.7 & 19.6 & 8.3\\
\midrule
\texttt{SD} & CASD & 13.4 & 25.2 & 17.0 & 8.3 &	14.9 & 21.0 & 10.0 & 5.9 & 0.5 & 18.0 &	12.8 \\
\midrule
\multirow{3}{*}{${\DFOSDA}_\text{casd}$} & \textbf{+FE} & 15.8 & 26.7 & 22.1 & 14.3 & 16.6 & 20.2 & 5.4 & 9.5 & 5.8 &  20.6	& 17.1 \\
\cline{2-13}
& \textbf{+OP} & 25.1 & 44.5 & 35.0 & 14.6 & 28.7 &	46.2 & 8.6 & 8.5 & 3.8 & 31.7 & 29.1\\
\cline{2-13}
& \textbf{+FE+OP} & 26.2 & 51.9 & 35.5 & 18.9 & 33.1 & 47.4 & 11.2 & 9.6 & 5.4 & 29.3 & 20.1 \\
\bottomrule
\end{tabular}
}

\end{subtable}

\end{table*}

\section{Additional Qualitative Analysis}

\begin{figure*}[hbt!]
  \centering
  \includegraphics[trim={0cm 0cm 0cm 0cm},clip,scale=0.5]{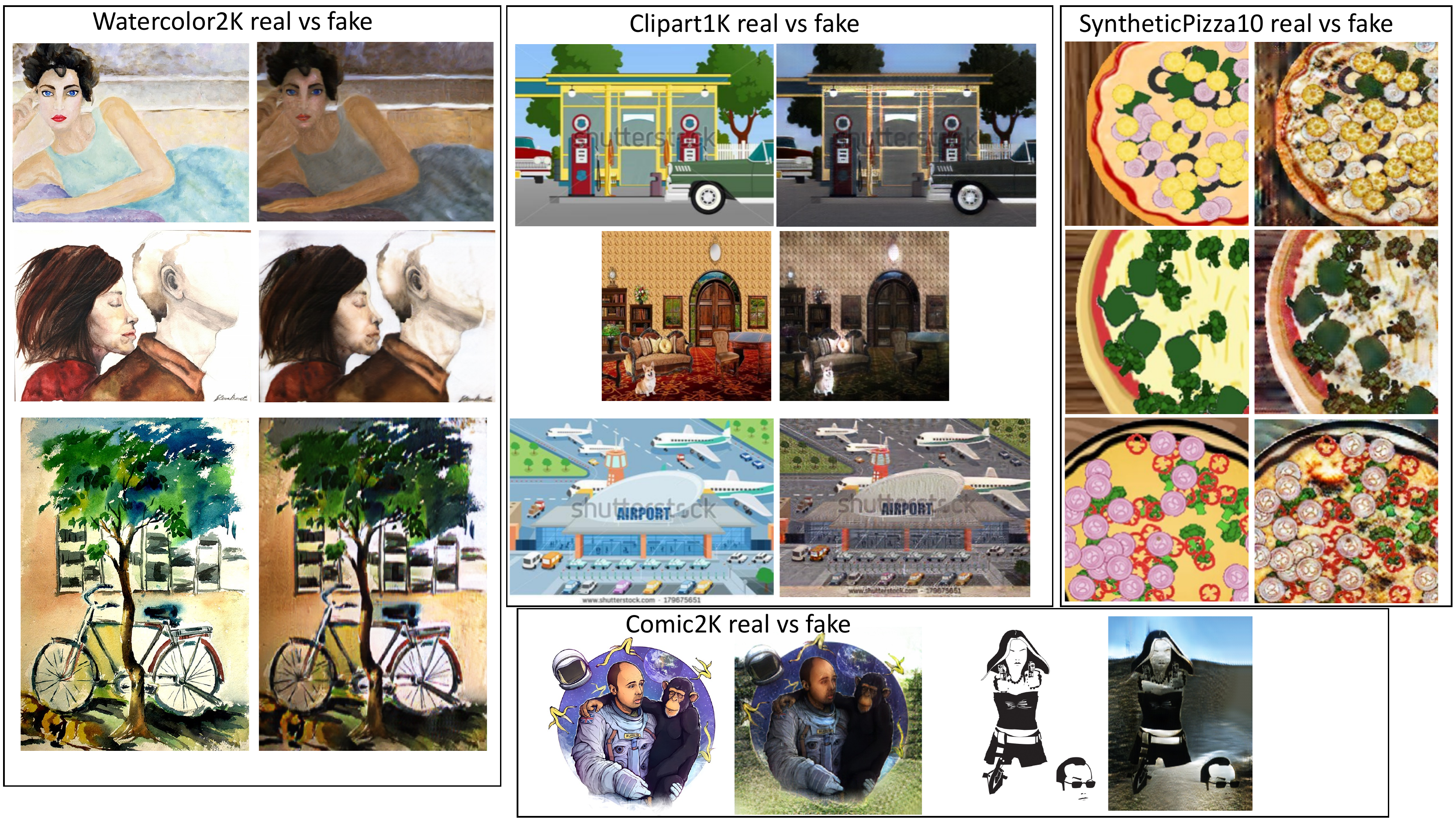}
  \vspace{0mm}
   \caption{Representative images generated by CycleGAN.
   }
   \label{fig:fsod2}
   \vspace{-3mm}
\end{figure*}

\autoref{fig:fsod2} shows the representative images generated by CycleGAN on different benchmarks.

\begin{figure*}[hbt!]
  \centering
  \includegraphics[trim={0cm 0cm +5cm 0cm},clip,scale=0.6]{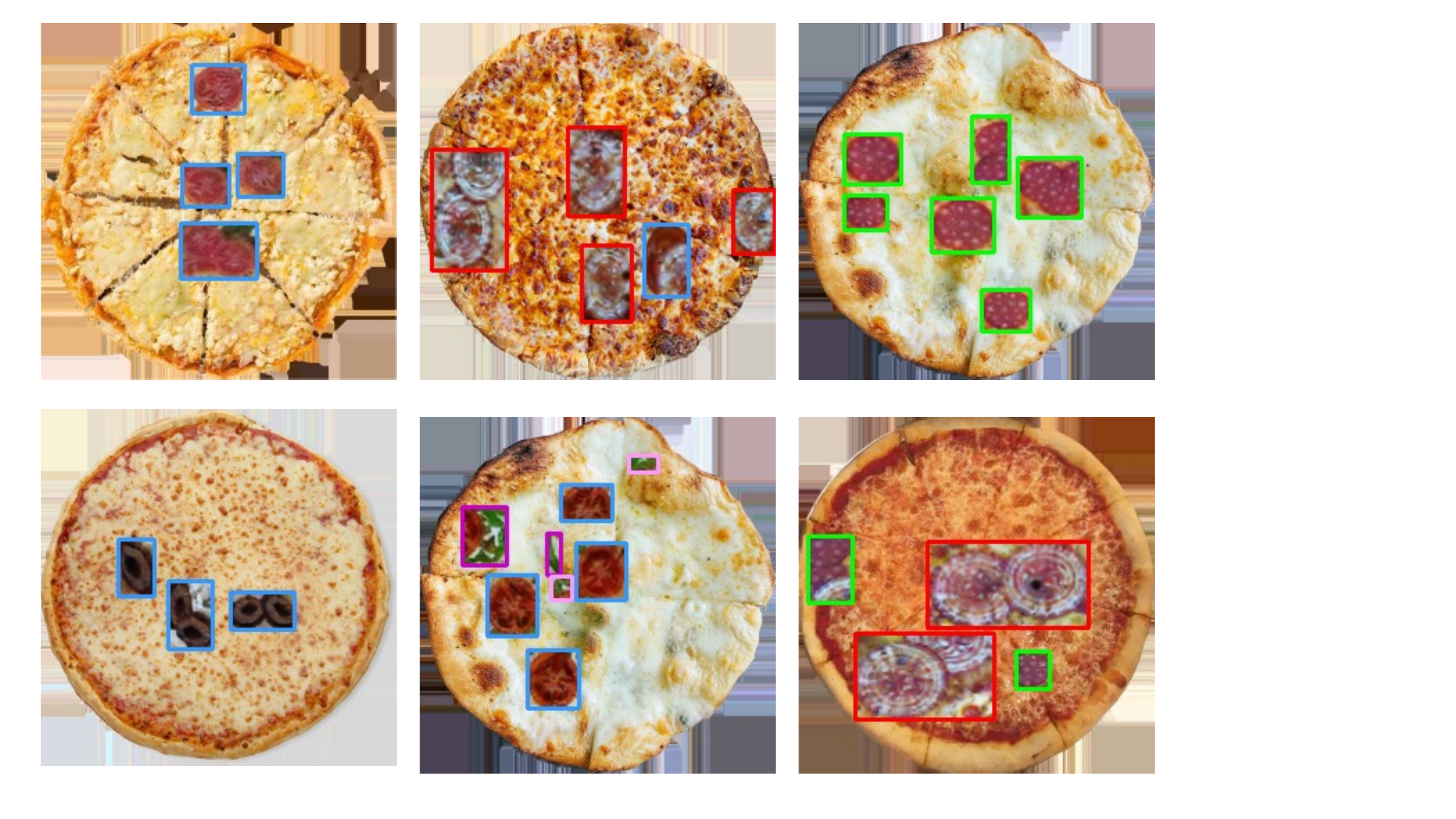}
  \vspace{-3mm}
   \caption{Representative images used for FSOD-3 training.
   }
   \label{fig:fsod3}
   \vspace{-3mm}
\end{figure*}

\autoref{fig:fsod3} illustrates some examples used for FSOD-3 training on SyntheticPizza10 $\rightarrow$ RealPizza10.

\begin{figure*}[hbt!]
  \centering
  \includegraphics[trim={0cm 0cm +7cm 0cm},clip,scale=0.6]{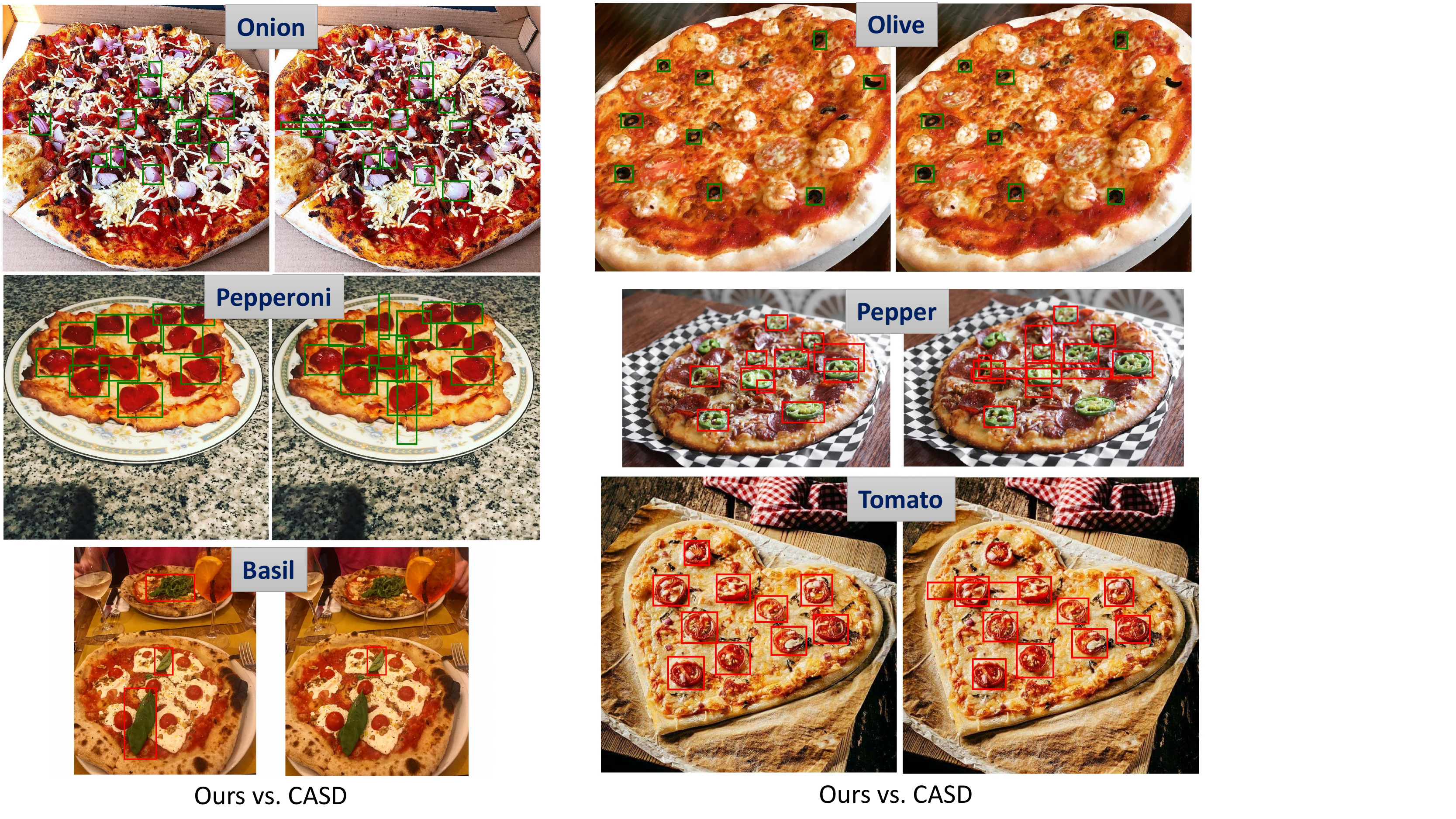}
  \vspace{-3mm}
   \caption{Example of success cases for our ${\DFOSDA}_\text{casd}$ vs. CASD in the test set of RealPizza10 dataset.
   We only show instances with scores over 0.3 to maintain visibility.
   }
   \label{fig:positives_supp}
   \vspace{-3mm}
\end{figure*}

\begin{figure*}[hbt!]
  \centering
  \includegraphics[scale=0.5]{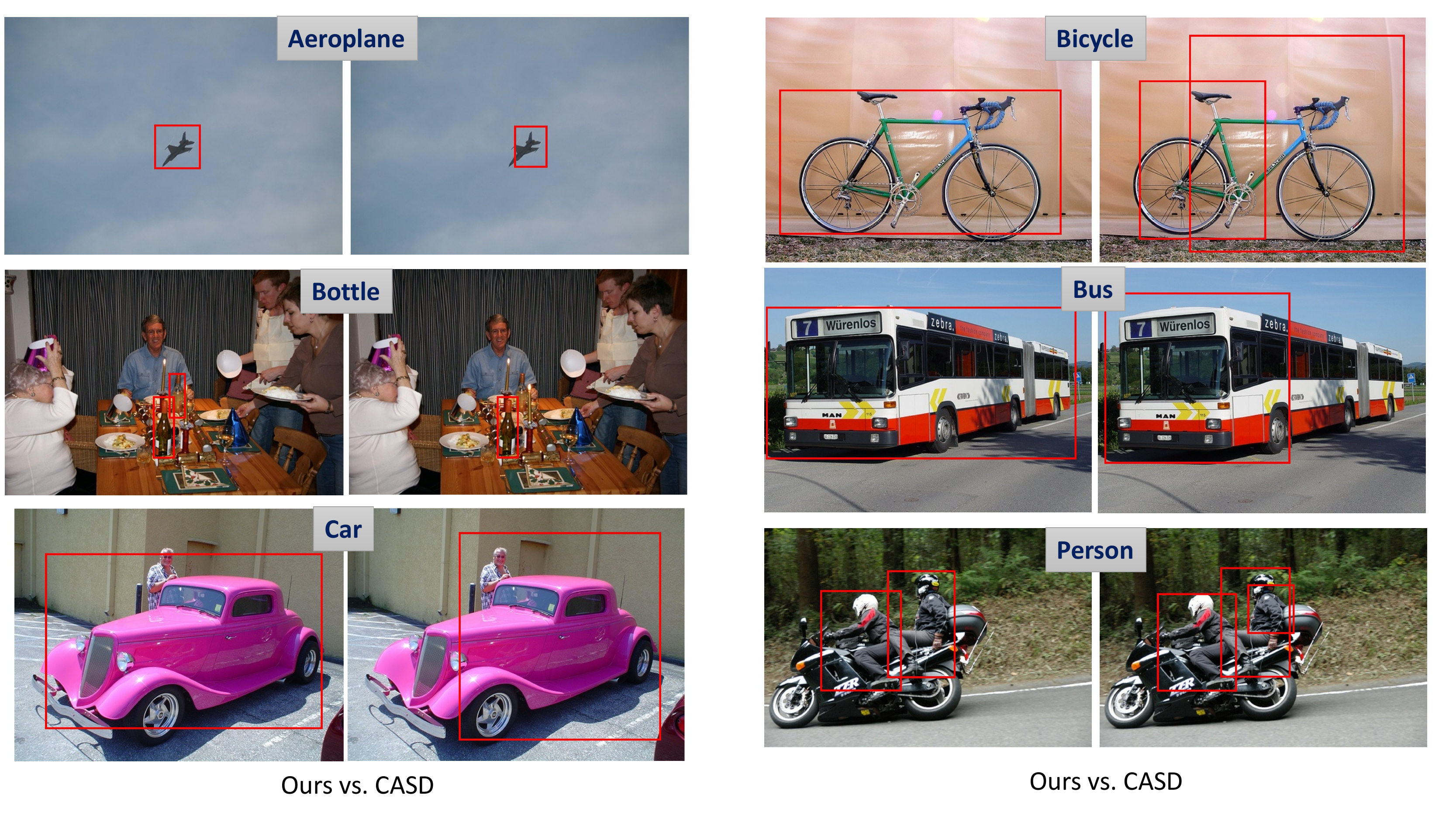}
  \vspace{-3mm}
   \caption{Example of success cases for our ${\DFOSDA}_\text{casd}$ vs. CASD in the test set of VOC2007 dataset.
   We only show instances with scores over 0.3 to maintain visibility.
   }
   \label{fig:positives_supp_voc}
   \vspace{-3mm}
\end{figure*}

\autoref{fig:positives_supp} and \autoref{fig:positives_supp_voc} illustrate  detection results produced by our \DFOSDA and CASD on RealPizza10 and VOC2007 datasets, respectively. There, it can be observed that \DFOSDA does not only locate most objects, but that it also produces more accurate bounding boxes. Specifically, in the RealPizza10 images it can be appreciated bounding boxes provided by our method (left) closely align with the objects of interest, while for CASD (right) bounding boxes are often imprecise (either wrong shape or big/small). Similar observations can be made for VOC2007 where CASD often fails to locate objects or produces spurious bounding boxes.

\begin{figure*}[hbt!]
  \centering
  \includegraphics[trim={0cm 0cm +8cm 0cm},clip,scale=0.5]{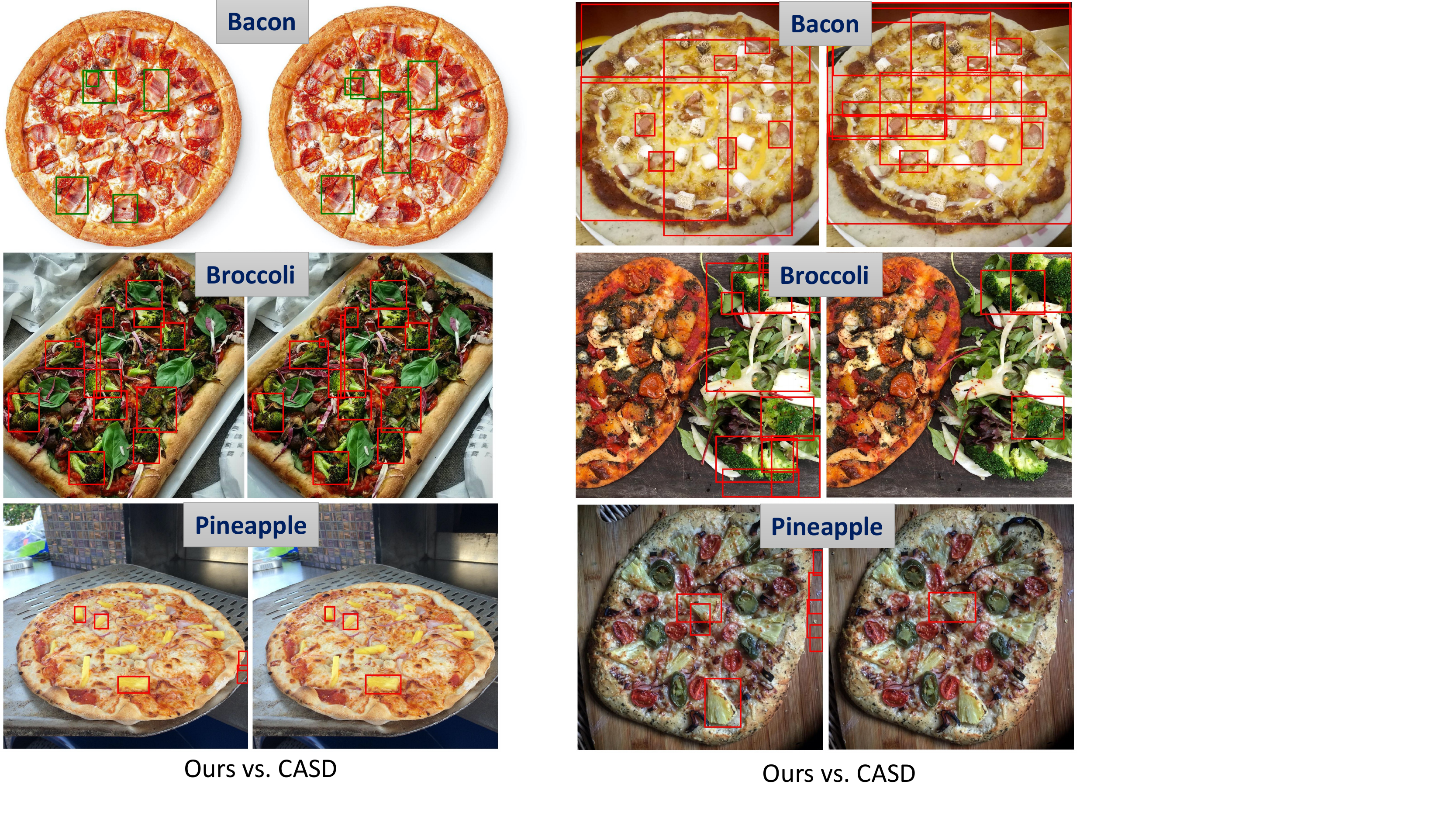}
  \vspace{-3mm}
   \caption{Challenging cases in the test set of Realpizza10 dataset where both our ${\DFOSDA}_\text{casd}$ and the baseline CASD fail. We hypothesize this is because these three categories have significantly smaller number of training examples.
   We only show instances with scores over 0.3 to maintain visibility.
   }
   \label{fig:positives_supp_failure}
   \vspace{-3mm}
\end{figure*}
\autoref{fig:positives_supp_failure} illustrates some challenging cases on the RealPizza10 dataset where the performance of both our \DFOSDA and the baseline CASD still lags. We hypothesize this is because in these cases, the corresponding categories (e.g., bacon, broccoli and pineapple) have significantly smaller number of training examples.

\end{document}